\documentclass[times,review,10pt]{elsarticle}

\usepackage{setspace}
\usepackage{etoolbox}

\AtBeginEnvironment{table}{\singlespacing}
\AtBeginEnvironment{table*}{\singlespacing}

\usepackage{amssymb}
\usepackage{amsmath}

\usepackage{xcolor}

\usepackage{hyperref}
\usepackage{threeparttable}
\usepackage{booktabs}
\usepackage{pdflscape}
\usepackage{rotating}

\usepackage{amsfonts}
\usepackage{graphicx}
\usepackage{textcomp}
\usepackage{multirow} 
\usepackage{bbding}

\usepackage{algorithm}
\usepackage[noend]{algpseudocode}

\journal{Pattern Recognition}

\begin{document}
\begin{frontmatter}



\title{DCMIL: Progressive Representation Learning of Whole Slide Images for Cancer Prognosis Analysis}


\author[1,2]{Chao Tu}
\author[1,3]{Ziying Wang}
\author[1,3]{Feng Li}
\author[4,5]{Kun Huang}
\author[6]{Jie Zhang}
\author[1,3]{Qianjin Feng}
\author[1,3]{Yu Zhang\corref{cor1}}
\author[1,3]{Zhenyuan Ning\corref{cor1}}

\affiliation[1]{organization={School of Biomedical Engineering},
            addressline={Southern Medical University}, 
            city={Guangzhou},
            postcode={510515}, 
            state={Guangdong},
            country={China}}
\affiliation[2]{organization={Department of Pathology and Institute of Molecular Pathology},
            addressline={The First Affiliated Hospital of Nanchang University}, 
            city={Nanchang},
            postcode={330006}, 
            state={Jiangxi},
            country={China}}
\affiliation[3]{organization={Guangdong Provincial Key Laboratory of Medical Image
Processing},
            addressline={Southern Medical University}, 
            city={Guangzhou},
            postcode={510515}, 
            state={Guangdong},
            country={China}}
\affiliation[4]{organization={Department of Biostatistics and Health Data Science},
            addressline={School of Medicine, Indiana University}, 
            city={Indianapolis},
            postcode={46202}, 
            state={IN},
            country={USA}}
\affiliation[5]{organization={Center for Computational Biology and Bioinformatics},
            addressline={School of Medicine, Indiana University}, 
            city={Indianapolis},
            postcode={46202}, 
            state={IN},
            country={USA}}
\affiliation[6]{organization={Department of Medical and Molecular Genetics},
            addressline={School of Medicine, Indiana University}, 
            city={Indianapolis},
            postcode={46202}, 
            state={IN},
            country={USA}}

\cortext[cor1]{Corresponding authors: jonnyning@foxmail.com (ZY. Ning) and yuzhang@smu.edu.cn (Y. Zhang).}

\begin{abstract}
The burgeoning discipline of computational pathology shows promise in harnessing whole slide images (WSIs) to quantify morphological heterogeneity and develop objective prognostic models for human cancers. However, progress is impeded by the computational bottleneck of gigapixel-size inputs and the scarcity of dense manual annotations. Current methods often overlook fine-grained information across multi-magnification WSIs and variations in tumor microenvironments. Here, we propose an easy-to-hard progressive representation learning framework, termed dual-curriculum contrastive multi-instance learning (DCMIL), to efficiently process WSIs for cancer prognosis. The model does not rely on dense annotations and enables the direct transformation of gigapixel-size WSIs into outcome predictions.
{Extensive experiments on twelve cancer types (5,709 patients and 12.54 million tiles) demonstrate consistently strong prognostic performance of DCMIL
compared with standard WSI-based prognostic models.}
Additionally, DCMIL identifies fine-grained prognosis-salient regions, provides robust instance uncertainty estimation, and captures morphological differences between normal and tumor tissues. All codes have been made publicly accessible at \href{https://github.com/tuuuc/DCMIL}{https://github.com/tuuuc/DCMIL}.
\end{abstract}

\begin{graphicalabstract}
\begin{center}
\includegraphics[width=\textwidth]{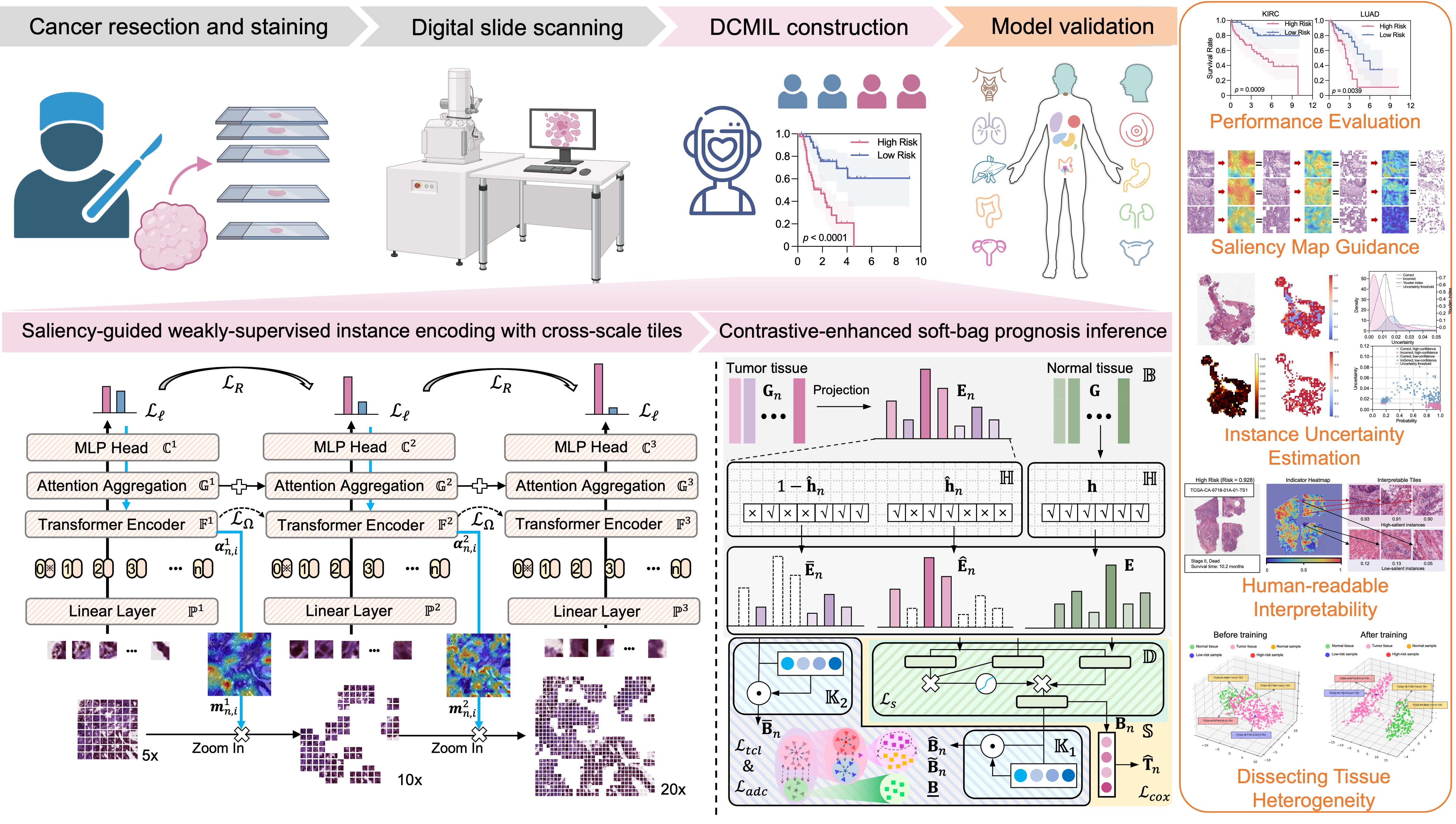}
\end{center}
\end{graphicalabstract}

\begin{highlights}
{
\item DCMIL extends dual-curriculum MIL with normal-control-aware triple-tier contrastive Cox learning.
\item Triple-tier contrast models tumor–normal, risk-level, and intra-bag saliency relationships.
\item Multi-magnification saliency guidance captures fine-grained prognosis-relevant WSI features.
\item Tumor–normal discrepancy and instance uncertainty provide interpretable model evidence.
\item Evaluation across 12 TCGA cancers supports robust prognostic performance.
}

\end{highlights}


\end{frontmatter}


\section{Introduction}
\label{sec:introduction}
Cancer, as a highly complex disease, exhibits significant heterogeneity in the tumor microenvironment that dominates the variability in patient outcomes~\cite{vitale2021intratumoral}.
Cancer prognosis analysis aims to stratify patients into different risk subgroups based on quantifying the heterogeneous characteristics of tumor~\cite{keyl2025decoding}.
Precise prognosis estimation can aid physicians in decision-making to optimize treatment strategies and improve patients' quality of life~\cite{alowais2023revolutionizing}.
Clinically, a general paradigm involves manual assessment of certain criteria (e.g., tumor stage) to evaluate patient prognosis based on histopathological features (e.g., tumor invasion, necrosis, and mitosis) or biomarkers (e.g., the morphology and location of tumor cells) driven from the histopathological slides~\cite{amin2017eighth}.
However, subjective evaluation may be subject to high inter-observer and intra-observer variability, and the identification of histopathological features or biomarkers typically requires labor-intensive efforts from pathologists~\cite{yan2025pathorchestra}.
Moreover, substantial variation in prognosis persists among patients, regardless of their classification into the same grade or stage~\cite{chen2022pan}.
Therefore, developing an objective, accurate, and robust prognostic model in this field remains critical.

Recently, high-resolution whole slide image (WSI) using digital slide scanners has triggered tremendous excitement for the burgeoning field of computational pathology~\cite{xu2024whole}.
Exploring the rich morphological information from WSIs, computational pathology has demonstrated promise in quantifying histopathological heterogeneity of the tumor microenvironment~\cite{zhang2026deep} and is competent in many tasks towards precision medicine (e.g., cancer diagnosis~\cite{zheng2025graphmamba}, biomarker discovery~\cite{zhou2026dual}, therapy and drug development~\cite{hoang2024deep}, and outcome prediction~\cite{neidlinger2025benchmarking}).
The emergence of artificial intelligence technology, especially deep learning approach, has further advanced computational pathology, which revolutionized the routine WSI analysis paradigm~\cite{zhang2025artificial}.
Current deep learning methods have been successfully applied to cancer prognosis analysis by learning latent prognosis-relevant morphological representations from WSIs~\cite{han2025inter}.
However, those methods have been impeded due to the computational bottleneck of gigapixel-size WSIs and the scarcity of manual dense annotations.
Although some studies try to alleviate this issue by taking small-sized region of interest (ROI) selectively sampled from WSIs as model's input, such scheme is experience-dependent and requires prior knowledge from pathologists~\cite{lu2021data}. 
Also, how to manually define prognosis-related ROIs is still an open issue, as such regions are generally distributed throughout the tumor and its microenvironment.

Multi-instance learning (MIL) method provides a feasible solution to perform inference in a weakly-supervised manner~\cite{zhang2026prioritized}, which has attracted increasing attention for WSI analysis in cancer prognosis estimation~\cite{tang2025intervention}.
MIL method adopts a divide-and-conquer strategy and typically consists of two stages, i.e., instance encoding stage and instance aggregation stage~\cite{shao2025do}.
In the instance encoding stage, previous methods have utilized pre-trained or foundation models to extract instance representations in an unsupervised or self-supervision way (i.e., without the reference of outcome labels)~\cite{luo2026ensemble}.
However, recent studies have demonstrated that the pre-trained model may not generalize well to the downstream tasks, as it is task-agnostic and inclined to overfit the pretraining objective~\cite{srinidhi2022self}.
Alternatively, some work has directly assigned each instance the bag-level annotation (i.e., the survival time of patient) for network training, which introduces task-specific supervision and improves model’s generalizability~\cite{vu2020novel}. 
Actually, the prognosis evaluation is primarily and comprehensively determined by certain representative regions, yet those regions might only occupy a small portion of WSI~\cite{zhang2022dtfd}. 
As a result, the strong supervision on all instances, especially on prognosis-irrelevant tiles, will introduce excessive label noise and limit model's performance. 
To alleviate this issue, a feasible way is to supervise the instance encoding in a weak and easy-to-hard manner~\cite{yang2022micl}.
The second stage of instance aggregation generally involves aggregating instance representations within a bag for prognosis estimation by certain fusion strategies (e.g., pooling operation and attention mechanism)~\cite{naik2020deep}.
Although such strategies are simple and straightforward, they may overwhelm prognosis-relevant information, cause intra-bag redundancy, and reduce inter-bag discrimination, when many instances from irrelevant regions are enrolled~\cite{wang2019rmdl}.
Additionally, the above-mentioned MIL methods primarily focus on single-scale local instance representations, 
such as the morphological changes or different growth patterns of tumor cells.
However, pathologists typically identify salient regions at a low-magnification field of view, and then transition to a higher-magnification level to examine fine-grained histopathological features within the context and assess differences in tumor microenvironments to discern better the nature and extent of cancer cells~\cite{song2023artificial}.


To address these challenges, we 
propose a progressive representation learning called Dual-Curriculum Contrastive Multi-Instance Learning (DCMIL) for efficient processing of whole slide images (WSIs) in cancer prognosis analysis. DCMIL incorporates two curricula designed to learn robust representations in an easy-to-hard manner. 
For the first curriculum, DCMIL introduces prognosis information by assigning risk stratification status (derived from survival time) as instance labels to learn instance-level representations in a weakly-supervised manner, which helps reduce label noise and maintain prognosis-related guidance.
Also, to imitate the reviewing procedure of pathologists, DCMIL leverages the low-magnification saliency map to guide the encoding of high-magnification instances for exploring fine-grained information across multi-magnification WSIs.
For the second curriculum, instead of enrolling all instances, DCMIL adaptively identifies and integrates representative instances within a bag (as the soft-bag) for prognosis inference. It leverages the constrained self-attention strategy to obtain additional sparsity for soft-bag representations, which can help reduce intra-bag redundancy at both instance and feature levels. 
Furthermore, contrastive learning is a self-supervised technique that learns effective representations by contrasting positive and negative samples and has been proven effective in various machine learning tasks~\cite{lu2024visual}. Consequently, DCMIL arms with triple-tier contrastive learning strategy for enhancing intra-bag (i.e., high-salient and low-salient regions within a subject) and inter-bag (i.e., \textcircled{1} tumor and normal tissues \textcircled{2} high-risk and low-risk regions between different subjects) discrimination. 
{Extensive experiments across twelve cancer types demonstrate consistently strong prognostic performance of DCMIL compared with diverse WSI-based benchmarks.}
Furthermore, DCMIL provides fine-grained prognosis-salient regions on WSIs, presents robust instance uncertainty estimation, and captures the morphological differences between tumor and normal tissue microenvironments, with the potential to generate new biological insights.
{ Recent studies have explored different strategies to enhance representation learning in weakly supervised WSI analysis. For example, CaMIL~\cite{chen2024camil} employs causal intervention to alleviate spurious associations, while CAMIL~\cite{fourkioti2024camil} introduces neighbor-constrained attention to capture spatial contextual information among instances. Additionally, PAMIL~\cite{he2025pseudo} leverages pseudo-label-guided attention to identify informative instances and refine feature aggregation. These approaches primarily focus on classification-oriented representation learning. By comparison, DCMIL adopts a prognosis-oriented progressive learning paradigm, integrating multi-magnification saliency guidance, adaptive soft-bag aggregation, and contrastive Cox optimization for survival prediction.
}

A preliminary version of this study was presented at NeurIPS 2022~\cite{tu2022dual}. 
That work established the core dual-curriculum MIL paradigm, including multi-magnification saliency-guided instance representation learning and soft-bag-based prognosis inference. Building on this framework, the present study further extends DCMIL in both methodology and experimental evaluation. A central extension to the contrastive survival-learning objective is the incorporation of adjacent-normal tissue controls into the dual-curriculum framework, yielding a normal-control-aware triple-tier contrastive Cox objective. Specifically, the risk-level inter-bag and saliency-based intra-bag relationships from the preliminary formulation are retained, while tumor--normal discrimination is introduced as an additional complementary contrastive tier using adjacent-normal tissue controls. This extension enables prognosis-related representations to be jointly regularized across risk-level, intra-bag saliency, and tumor--normal relationships within a unified survival-learning framework. In addition, the journal version substantially expands the evaluation and characterization of DCMIL. We incorporate instance-level uncertainty and calibration analyses, controlled ablations of the contrastive objectives, hyperparameter and robustness analyses, evaluation across twelve TCGA cancer cohorts, and independent cross-cohort validation on CPTAC-HNSCC. Together, these extensions provide a more comprehensive assessment of the robustness, reliability, and applicability of the proposed framework.


\section{Method details}
\subsection{Study Population}
We developed and evaluated our proposed method on 12 datasets collected from The Cancer Genome Atlas (TCGA) database \footnote{https://portal.gdc.cancer.gov/}.
{After excluding patients with a survival time of less than one month from each dataset, 5,709 patients were retained for the final experiments,}
including Bladder Urothelial Carcinoma (BLCA, $N = 395$), Breast Invasive Carcinoma (BRCA, $N = 993$), Colon Adenocarcinoma (COAD, $N = 352$), Head and Neck Squamous Cell Carcinoma (HNSC, $N = 430$), Kidney Renal Clear Cell Carcinoma (KIRC, $N = 440$), Liver Hepatocellular Carcinoma (LIHC, $N = 333$), Lung Adenocarcinoma (LUAD, $N = 438$), Lung Squamous Cell Carcinoma (LUSC, $N = 425$), Ovarian Serous Cystadenocarcinoma (OV, $N = 528$), Stomach Adenocarcinoma (STAD, $N = 370$), Thyroid Carcinoma (THCA, $N = 501$), and Uterine Corpus Endometrial Carcinoma (UCEC, $N = 504$). 
Each dataset contained WSIs stained with hematoxylin and eosin (H\&E) along with clinical information (i.e., survival time and event status). 
After segmenting tissue areas using Otsu's method, a set of non-overlapping tiles at different magnifications ($20\times$, $10\times$, $5\times$) were sampled from each segmented tissue area, with window sizes set to $512\times512$, $256\times256$, and $128\times128$, respectively.
\subsection{Problem Formulation.} 
\begin{figure}[ht]
	\centering
	\includegraphics[width=0.9\textwidth]{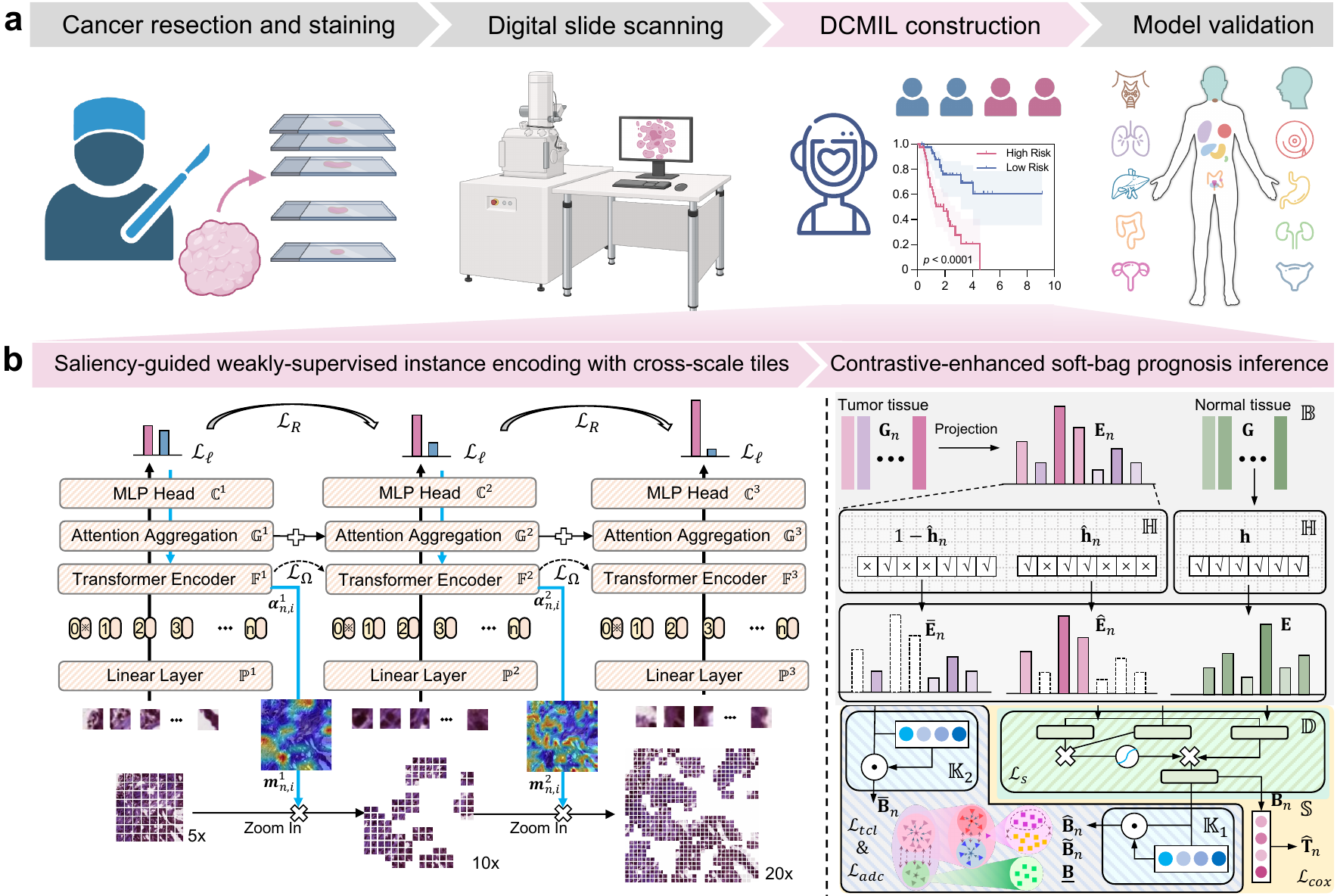}
	\caption{\textbf{Overview of the framework.} 
    $\textbf{a}$, The overall workflow, including cancer resection and staining, digital slide scanning, DCMIL construction, and model validation.
    $\textbf{b}$, The pipeline of DCMIL comprises two easy-to-hard curricula, including 1) saliency-guided weakly-supervised instance encoding with cross-scale tiles, and 2) contrastive-enhanced soft-bag prognosis inference.
	}
    \vspace{-10pt}
	\label{fig_1}
\end{figure}

As illustrated in Figure~\ref{fig_1}a, the overall workflow of this study consists of four major components: cancer tissue resection and staining, WSI scanning, DCMIL model construction, and model validation.
Figure~\ref{fig_1}b presents the detailed architecture of DCMIL, which adopts an easy-to-hard learning paradigm composed of two progressive curricula:
(1) saliency-guided weakly supervised instance encoding with cross-scale tiles, and
(2) contrastive-enhanced soft-bag prognosis inference.
The main notations used in this paper are summarized in Table~S8 (in \textit{Supplementary Materials}) for reference.
Denote the triplet $\{\mathbf X_n, \mathbf T_n,\boldsymbol{\delta}_n\}^N_{n=1}$ as the dataset consisting of $N$ patients, where $\mathbf X_n$ is the $n$-th patient with one or more WSIs, and $\mathbf T_n$ and $\boldsymbol{\delta}_n$ are respectively its observed survival time and event indicator (i.e., equal to 1 and 0 for uncensored and censored patients, respectively). 
Prognosis analysis is an ordinal regression task that models time-to-event distribution, which is formulated as
{ \small \begin{equation} 	
	\widehat{\mathbf T}_n=\mathbb{S}(\mathbf X_n),
\end{equation}}
where $\mathbb{S}$ and $\widehat{\mathbf T}_n$ denote the prognosis inference function and the predicted prognostic risk score, respectively.
In practice, it is difficult to directly feed the WSI into the model due to its gigapixel size. 
Accordingly, many studies \cite{chen2021whole} have decomposed the WSI into a large number of tiles and generated a bag $\mathbf X_n=\{\mathbf x_{n,i}\}_{i=1}^{N_n}$ that contains $N_n$ instances (tiles) for the $n$-th patient. 
Then, MIL is leveraged to construct the prognosis model as follows:
{ \small \begin{equation} 	
	\widehat{\mathbf T}_n=\mathbb{S}(\mathbb{A} \{\mathbb{E}(\mathbf x_{n,i}):\mathbf x_{n,i}\in{\mathbf X_n}\}),
	\label{mil}
\end{equation}}
where $\mathbb{E}$ is an instance encoding function that extracts feature representation for each instance, and $\mathbb{A}$ is a permutation-invariant instance aggregation function that pools the instance representations into the bag ones.
However, previous studies \cite{zhang2022dtfd} typically generate instance representations via a pre-trained model or a model trained by the instances with bag-level annotations, which may not generalize well to the downstream task. 
As a result, we decompose the model into two easy-to-hard curricula, including (1) Curriculum I (C-I): instance encoding (via ${\mathbb{E}}$) during preliminary risk stratification (via ${\mathbb{C}}$, a risk stratification function), and (2) Curriculum II (C-II): prognosis inference (via ${\mathbb{S}}$) after instance aggregation (via ${\mathbb{A}}$).
Formally, the optimization process can be formulated as
{ \small \begin{equation}  
	\left\{  
	\begin{aligned} 
		&\text{C-I}: \widehat{\mathbb{E}}, \widehat{\mathbb{C}}=\arg \min_{\mathbb{E},\mathbb{C}} \sum_{n=1}^N\dag(\mathbf Y_n==1\|0)\sum_{i=1}^{N_n}[\mathbb{C}\{\mathbb{E}(\mathbf x_{n,i}):\mathbf x_{n,i}\in{\mathbf X_n}\}, \mathbf Y_n]_{\mathcal L_{\textrm{I}}}\\
		&\text{C-II}: \widehat{\mathbb{S}}, \widehat{\mathbb{A}}= \arg \min_{\mathbb{S},\mathbb{A}} \sum_{n=1}^N[\mathbb{S}(\mathbb{A}\{\widehat{\mathbb{E}}(\mathbf x_{n,i}):\mathbf x_{n,i}\in{\mathbf X_n}\}), \mathbf T_n, \boldsymbol{\delta}_n]_{\mathcal L_{\textrm{II}}}
	\end{aligned}
	\right.,
	\label{targe}
\end{equation}} 
where $\dag(\cdot)$ outputs 1 (0) if true (false), and ${\mathcal L_{\textrm{I}}}$ and ${\mathcal L_{\textrm{II}}}$ denote the loss functions for C-I and C-II, respectively.
The risk stratification status $\mathbf Y_n$ is determined by $\{\mathbf T_n, \boldsymbol{\delta}_n\}$ with a three-year (or five-year) time threshold (denoted as $\mathbf T_r$) as follows:
{ \small \begin{equation}
	\mathbf Y_n=\left\{ 
	\begin{aligned}
		&1, &&\mathbf T_n\leq \mathbf T_r \ \&\&\ \boldsymbol{\delta}_n==1 \\ 
		&0, &&\mathbf T_n> \mathbf T_r \\
		&\rm -, && \rm{others}
	\end{aligned} 
	\right.
\end{equation}}

\subsection{Curriculum I: Saliency-guided weakly-supervised instance encoding with cross-scale tiles.}
\begin{figure}
	\centering
	\includegraphics[width=0.95\textwidth]{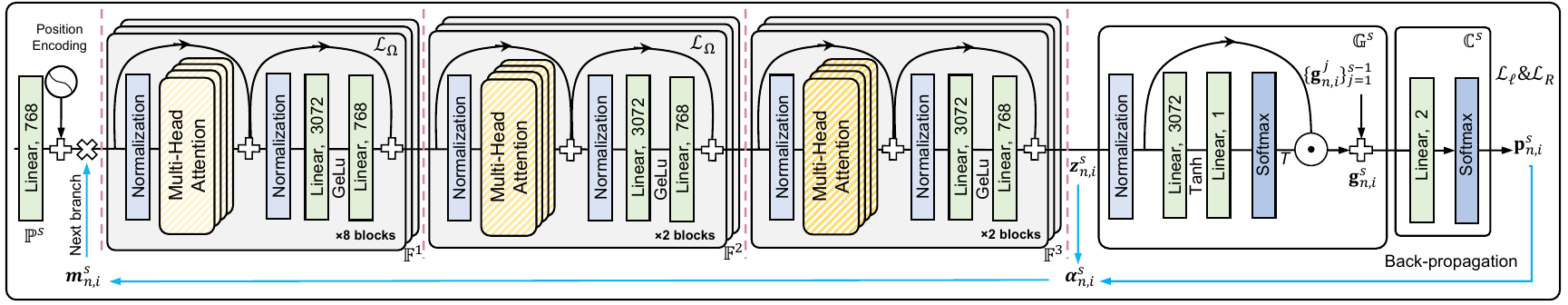}
	\caption{\textbf{The detailed architectures on Curriculum I.} For the $s$-th branch, its architecture mainly consists of a linear layer $\mathbb{P}^s$, a transformer encoder $\mathbb{F}^s$, an attention aggregator $\mathbb{G}^s$ and a multilayer perceptron (MLP) head $\mathbb{C}^s$.
	}
	\label{fig_1.1}
    \vspace{-10pt}
\end{figure}
The pipeline of Curriculum I is shown in Figure~\ref{fig_1}b.
Given the input instance with $S$ magnifications $\{\mathbf x_{n,i}^s\}_{s=1}^S$ ($S=3$ in this work), the proposed method contains $S$ branches and each branch takes different-magnification tiles as input, which aims to explore multi-scale information from multi-magnification images.
For the $s$-th branch, it mainly consists of a linear layer $\mathbb{P}^s$, a transformer encoder $\mathbb{F}^s$, an attention aggregator $\mathbb{G}^s$ and a multilayer perceptron (MLP) head $\mathbb{C}^s$. 
$\{\mathbb{P}^s\}_{s=1}^S$, $\{\mathbb{F}^s\}_{s=1}^S$, and $\{\mathbb{G}^s\}_{s=1}^S$ form $\mathbb{E}$, while $\{\mathbb{C}^s\}_{s=1}^S$ constitutes $\mathbb{C}$ in Eq.(\ref{targe}).
Specifically, the input instance (i.e., tile) is first divided into several $16\times16$ tokens. 
As exhibited in Figure~\ref{fig_1.1}, $\mathbb{P}^s$ is used to map the flattened tokens into 768-dimensional tokens. Following position encoding, $\mathbb{F}^s$ is placed to extract token representation, which is a ViT-like network architecture with multiple blocks.
Each block contains a multi-head attention layer, an MLP, and two normalization layers. The MLP comprises two linear layers with 3072 and 768 nodes, respectively, in which "GeLu" activation function is inserted between the two linear layers. 
Two normalization layers are respectively positioned before the multi-head attention layer and the MLP.
Next, $\mathbb{G}$ processes all tokens through a normalization layer, an MLP, and a "Softmax" layer to obtain token attention. The MLP comprises two linear layers with 3072 and 1 nodes, respectively, in which "Tanh" activation function is inserted between the two linear layers. The tokens are then aggregated based on token attention, and the tile representation from the last scale is summarized through a skip connection, yielding the tile representation for this scale. 
Finally, $\mathbb{C}$ predicts the risk stratification status using a linear layer followed by a "Softmax" layer.

Different from previous studies that separately feed $\{\mathbf x_{n,i}^s\}_{s=1}^S$ into the network and ignore fine-grained details across multi-magnification images, we encourage the network to focus on saliency regions by utilizing the prior knowledge of the ($s-1$)-th branch to highlight the input $\mathbf x_{n,i}^s$, which can be formulated as
{ \small \begin{equation}
    \widehat{\mathbf x}_{n,i}^s=\left\{ \begin{aligned}
    & \mathbf x_{n,i}^s&&,s=1 \\ 
    & {\mathbf m_{n,i}^{s-1}}\otimes\mathbf x_{n,i}^s&&,s>1 \\ 
    \end{aligned} \right.,
    \label{input}
\end{equation}} 
where $\otimes$ denotes the element-wise multiplication operator, and ${\mathbf m_{n,i}^{s-1}}$ 
\footnote{{Notably, the saliency map generated from the previous branch is detached from the computational graph before being used as a guidance mask for the next branch, which means that the GradCAM-based saliency masking in Eq. (5) does not introduce gradient propagation across branches during training.}}
is the saliency mask indicating high-risk regions, computed by gradient class activation map (GradCAM).
For convenience, we introduce how to acquire ${\mathbf m_{n,i}^{s}}$, which is easily generalized to ${\mathbf m_{n,i}^{s-1}}$.
Specifically, for each branch, we first generate an instance token embedding $\mathbf z_{n,i}^{s}$ through $\mathbb{P}^s$ and $\mathbb{F}^s$:
{ \small \begin{equation}
    \mathbf z_{n,i}^{s}=\mathbb{F}^{s}(\mathbb{P}^{s}(\widehat{\mathbf x}_{n,i}^{s})).
\end{equation}}  
Then, we aggregate $\mathbf z_{n,i}^{s}$ to obtain the instance representation $\mathbf g_{n,i}^s$ via $\mathbb{G}^s$, while assimilating comprehensive information from the current and preceding branches, which is computed by
{ \small \begin{equation}
    {\mathbf g_{n,i}^s=\mathbb{G}^{s}(\mathbf z_{n,i}^s, \mathbf g_{n,i}^{s-1}).}
    \label{g}
\end{equation}}
And $\mathbf g_{n,i}^s$ is used to predict the risk stratification status through $\mathbb{C}^s$:
{ \small \begin{equation}
	\mathbf p_{n,i}^{s}=\mathbb{C}^{s}(\mathbf g_{n,i}^s),
\end{equation}}
where $\mathbf p_{n,i}^{s}$ denotes the probability of risk stratification status.
After completing the forward propagation, we can obtain the gradient of each token through backward propagation, thereby generating the saliency mask ${\mathbf m_{n,i}^{s}}$:
{ \small \begin{equation}
	\alpha^s_{n,i} = \frac{\partial {\mathbf p_{n,i}^{s}}}{\partial \mathbf z_{n,i}^{s}},
\end{equation}}
{ \small \begin{equation}
	{\mathbf m_{n,i}^{s}}= \mathbb{R}(\alpha^s_{n,i} \mathbf z_{n,i}^{s})\geq \iota,
	\label{m}
\end{equation}}
where $\alpha_{n,i}^s$ is the gradient of the score for $\mathbf p_{n,i}^{s}$, $\mathbb{R}$ represents the ``ReLU” activation function and $\iota$ is a threshold set at 0.4.

The empirical loss $\mathcal L_{\ell}$ for Curriculum I is defined as
{ \small \begin{equation}
	\mathcal L_{\ell}=-\frac{1}{NN_nS}\sum_{n=1}^{N}\sum_{i=1}^{N_n}{\sum_{s=1}^{S}}{\mathbf y_{n,i}^{s} \log(\mathbf p_{n,i}^{s})+(1-\mathbf y_{n,i}^{s}) \log(1-\mathbf p_{n,i}^{s})},
\end{equation}}
where $\mathbf y_{n,i}^{s} \in{\mathbf Y_n} $ denotes the risk stratification status of the $i$-th instance in the $n$-th bag. 
Furthermore, we introduce a hierarchical transfer strategy to leverage the prior from different-magnification images, accelerate network convergence, and maintain the specificity of the current branch. 
Specifically, as shown in Figure~\ref{fig_1}b, the parameter $\boldsymbol \theta^{s}_{\mathbb{F}^s}$ (for any $s$) of the $s$-th block in $\mathbb{F}^s$ is fully learnable and unconstrained, while {$\{\boldsymbol\theta^{i}_{\mathbb{F}^s}\}_{i=1}^{s-1}$} (for $s>1$) is also learnable but constrained to approximate {$\{\boldsymbol\theta^{i}_{\mathbb{F}^{s-1}}\}_{i=1}^{s-1}$} by the structural loss $\mathcal L_{\Omega}$ as follows:
{{ \small \begin{equation}
	\mathcal L_{\Omega}=\left\| \{\boldsymbol\theta^{i}_{\mathbb{F}^{s}}\}_{i=1}^{s-1}-\{\boldsymbol\theta^{i}_{\mathbb{F}^{s-1}}\}_{i=1}^{s-1} \right\|_2.
	\label{L2}
\end{equation}}
To encourage the finer-scale branch to take more confident prediction from the coarse-scale branch as references, we further introduce a pairwise ranking loss $\mathcal L_{R}$ that has the following definition:
{ \small \begin{equation}
	{ \mathcal L_{R}=\sum_{n=1}^{N}\sum_{i=1}^{N_n}\sum_{s=2}^{S} {\max (0,\eta-\mathbf y_{n,i}^{s} \log \frac {\mathbf p_{n,i}^{s}}{\mathbf p_{n,i}^{s-1}}-(1-\mathbf y_{n,i}^{s}) \log(\frac{1-\mathbf p_{n,i}^{s}}{1-\mathbf p_{n,i}^{s-1}}))},}
\end{equation}}
where $\eta$ is a hyperparameter set to $1e^{-3}$ to make the prediction in the finer-scale branch more confident. 
Intuitively, when the instance label is high-risk, the high-risk prediction probability in the fine-scale branch is greater than that in the coarse-scale branch. 
The same applies to low-risk instances.
Finally, a hybrid loss function $\mathcal L_{\textrm{I}}$ is designed to train the network, which contains three parts as follows:
{ \small \begin{equation}
	\mathcal L_{\textrm{I}}=\mathcal L_{\ell}+\beta_{\Omega} \mathcal L_{\Omega} {+\beta_{R} \mathcal L_{R}},
\end{equation}}
where $\beta_{\Omega}$ and $\beta_{R}$ are the weight coefficients.
After network optimization, we can obtain ${\mathbf g^S_{n,i}}$ as multi-scale instance representation ${\mathbf g_{n,i}}$ (i.e., ${\mathbf g_{n,i}} = \mathbf g^S_{n,i}$, as ${\mathbf g^S_{n,i}}$ encompasses comprehensive information from the current and preceding branches by Eq.(\ref{g})) for subsequent prognosis inference task.

\subsection{Curriculum II: Contrastive-enhanced soft-bag prognosis inference.}	
\label{c2}
\begin{figure}
	\centering
    \includegraphics[width=0.95\textwidth]{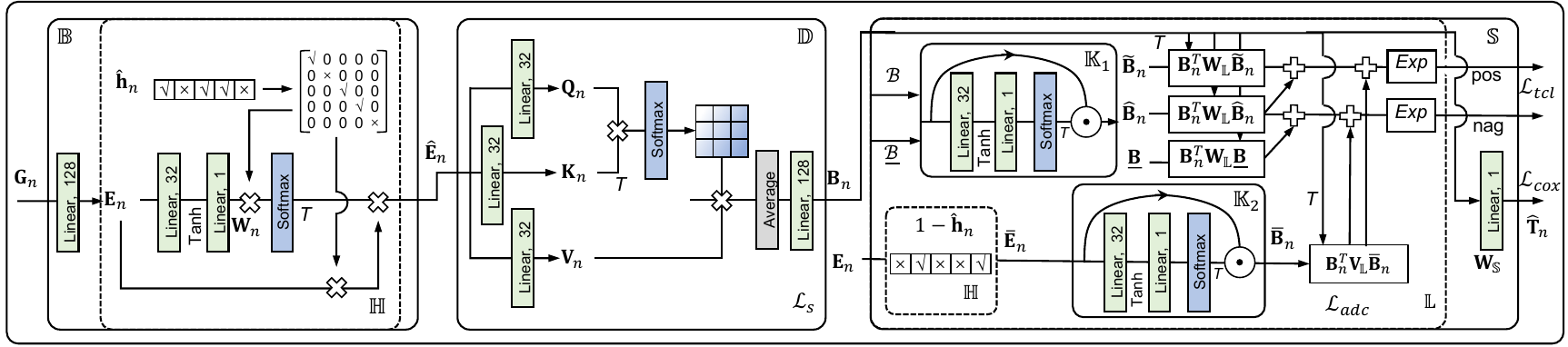}
	\caption{\textbf{The detailed architecture on Curriculum II}, containing a soft-bag learning module $\mathbb{B}$, a constrained self-attention module $\mathbb{D}$, and a contrastive-enhanced prognosis inference function $\mathbb{S}$.
	}
	\label{fig_1.2}
    \vspace{-10pt}
\end{figure}
In this section, we develop the second curriculum, and its pipeline is shown in Figure~\ref{fig_1}b.
Given a set of instance representations from the $n$-th bag, i.e., $\mathbf G_n=[\mathbf g_{n,1};\mathbf g_{n,2};\cdots;\mathbf g_{n, N_n}]^{T} \in{\mathbb{R}^{N_n\times C}}$, where $C$ denotes the feature dimension, the proposed method mainly consists of a soft-bag learning module $\mathbb{B}$, a constrained self-attention module $\mathbb{D}$, and a contrastive-enhanced prognosis inference function $\mathbb{S}$, where $\mathbb{B}$ and $\mathbb{D}$ form $\mathbb{A}$ in Eq.(\ref{targe}).

For $\mathbb{B}$ (as shown in Figure~\ref{fig_1.2}), it first learns new bag representation $\mathbf{E}_n \in{\mathbb{R}^{N_n\times D}}$ (where $D$ denotes the feature dimension) from $\mathbf{G}_n$ via a linear layer.
Instead of enrolling all instances, we adaptively identify and integrate representative instances within a bag by introducing the function $\mathbb{H}$:
{ \small \begin{equation}
	\widehat{\mathbf{E}}_n=\mathbb{H}(\mathbf{E}_n,\widehat{\mathbf{h}}_n)={\phi(diag\{\widehat{\mathbf{h}}_n\}\times\mathbf{W}_n)}^{T}\times(diag{\{\widehat{\mathbf{h}}_n\}\times\mathbf{E}_n}),
	\label{sb}
\end{equation}}
where $\widehat{\mathbf{E}}_n \in{\mathbb{R}^{N_B \times D}}$ ($N_B$ is the number of selected instances in each bag, and $N_B\ll N_n$) is the feature representation of those selected instances, {$\phi$ represents the “Softmax” activation function,} and $\mathbf{W}_n\in{\mathbb{R}^{N_n \times N_B}}$ denotes the weight matrix that is generated by feeding $\mathbf{E}_n$ into two linear layers (as shown in Figure~\ref{fig_1.2}).
And $\widehat{\mathbf{h}}_n\in{\mathbb{R}^{N_n\times 1}}$ is an indicator that adaptively selects $N_B$ representative instances and is computed by
{ \small \begin{equation}
	\widehat{\mathbf{h}}_n={\arg \max_{\mathbf{h}_n}}\ \mathbb{S}\left(\mathbb{D}\left(\mathbb{H}(\mathbf{E}_n,{\mathbf{h}}_n)\right)\right), \quad \ s.t.\sum_{i=1}^{N_n}{\mathbf{h}_{n,i}}=N_B \ \&\&\ \mathbf{h}_{n,i} \in \{0,1\}.
	\label{h}
\end{equation}}
Since the top-$N_B$ assignment operation via "$\arg \max$" is not differentiable, we instead use the Gumbel-Softmax strategy to compute $\widehat{\mathbf{h}}_n$.


Subsequently, we leverage $\mathbb{D}$ to obtain additional sparsity for soft-bag representations, as shown in Figure~\ref{fig_1.2}. 
Given the input $\widehat{\mathbf{E}}_n$, it first linearly projects $\widehat{\mathbf{E}}_n$ into three feature spaces (i.e., $\mathbf{Q}_n \in\mathbb{R}^{N_B \times \widehat{D}}$, $\mathbf{K}_n \in\mathbb{R}^{N_B \times \widehat{D}}$, and $\mathbf{V}_n \in\mathbb{R}^{N_B \times \widehat{D}}$, and $\widehat{D}$ is the feature dimension) via three mapping matrices (i.e., $\mathbf{W}_{Q} \in\mathbb{R}^{D \times \widehat{D}}$, $\mathbf{W}_{K} \in\mathbb{R}^{D \times \widehat{D}}$, and $\mathbf{W}_{V} \in\mathbb{R}^{D \times \widehat{D}}$), in which the mapping matrices are constrained by 
{ \small \begin{equation}
	\mathcal{L}_{s}=(\left\|\mathbf{W}_{Q}\right\|_1+\left\|\mathbf{W}_{K} \right\|_1+\left\|\mathbf{W}_{V} \right\|_1).
\end{equation}}
Then, the sparse soft-bag representation $\mathbf{B}_n \in {\mathbb{R}^{1\times D_B}}$ (where $D_B$ is the feature dimension) can be obtained by the correlation-based activation mechanism, as illustrated in Figure~\ref{fig_1.2}.
It is worth mentioning that the collaboration of $\mathbb{B}$ and $\mathbb{D}$ can significantly help reduce intra-bag redundancy at both instance and feature levels so as to improve model's generalization. 

For $\mathbb{S}$, it takes $\mathbf{B}_n$ as input to infer the relative risk $\widehat{\mathbf T}_n$ via a linear projection matrix $\mathbf{W}_\mathbb{S} \in \mathbb{R}^{D_B \times 1}$, as shown in Figure~\ref{fig_1.2}.
In order to boost the ability of soft-bag inference, we introduce both tumor tissue and normal samples with $\mathcal O$ and $\underline{\mathcal O}$ distributions, respectively, and perform triple-tier contrastive learning to enhance intra-bag and inter-bag discrimination.
We assume that a bag $\mathbf{B}_n$ is sampled from $\mathcal O=\{\mathcal O^+,\mathcal O^-\}$, where the positive source $\mathcal O^+$ denotes the distribution of high-risk samples (i.e., those with survival time $\leq \mathbf T_r$), while the negative source $\mathcal O^-$ presents the distribution of low-risk samples (i.e., those with survival time $> \mathbf T_r$). 
Additionally, we define $\mathcal B_n^{\mathcal O}$ as the collection of bags (excluding $\mathbf{B}_n$) from the source $\mathcal O$, and $\mathcal B^{\underline{\mathcal O}}$ as the one from the source $\underline{\mathcal O}$.
Accordingly, $\mathcal{B}_n^{\mathcal{O^+}}$ (or $\mathcal{B}_n^{\mathcal{O^-}}$) is defined as the collection of bags from ${\mathcal{O^+}}$ (or ${\mathcal{O^-}}$) with the same source as $\mathbf{B}_n$.
We first focus on inter-bag (i.e., \textcircled{1} tumor and normal tissues \textcircled{2} high-risk and low-risk subjects) discrimination. 
For the former, it is expected to distinguish the tumor tissue from normal tissue.
Specifically, we obtain the integrated representations of $\mathcal B_n^{\mathcal O}$ and $\mathcal B^{\underline{\mathcal O}}$ (denoted as $\widetilde{\mathbf{B}}_n$ and $\underline{\mathbf{B}}$\footnote{
{Normal tissue slides were TCGA adjacent-normal samples assigned by patient-level folds and used only as training-fold controls in Curriculum II, with test-fold normal tissue excluded from both training and testing}
}, respectively) by merging the representations of all bags in $\mathcal B_n^{\mathcal O}$ and {$\mathcal B^{\underline{\mathcal O}}$} through a {bag} aggregation network $\mathbb{K}_1$ (as illustrated in Figure~\ref{fig_1.2}). 
Inter-bag discrimination (towards tumor and normal tissues) can be enhanced by maximally preserving the mutual information between $\mathbf{B}_n$ and $\widetilde{\mathbf{B}}_n$ so as to "pick up" $\widetilde{\mathbf{B}}_n$ from normal tissues:
{ \small \begin{equation}
	\label{kkk2}
	\sum_{n\in[N]} \mathbb{I}(\mathbf{B}_n,\widetilde{\mathbf{B}}_n)=\sum_{n\in[N]}p(\mathbf{B}_n,\widetilde{\mathbf{B}}_n)\log\frac{p(\mathbf{B}_n|\widetilde{\mathbf{B}}_n)}{p(\mathbf{B}_n)}.
\end{equation}}
Similarly, we obtain the representation of $\mathcal{B}_n^{\mathcal{O^+}}$ {(or $\mathcal{B}_n^{\mathcal{O^-}}$)} (denoted as $\widehat{\mathbf{B}}_n$) via $\mathbb{K}_1$.
And inter-bag discrimination (towards high-risk and low-risk subjects) can be formulated as
{ \small \begin{equation}
	\label{kkk}
	\sum_{n\in[N]} \mathbb{I}(\mathbf{B}_n, \widehat{\mathbf{B}}_n)=\sum_{n\in[N]}p(\mathbf{B}_n,\widehat{\mathbf{B}}_n)\log\frac{p(\mathbf{B}_n|\widehat{\mathbf{B}}_n)}{p(\mathbf{B}_n)}.
\end{equation}}
Subsequently, we concentrate on intra-bag (i.e., high-salient and low-salient regions within a subject) discrimination.
According to Eq.(\ref{sb}), partial instances within a bag are adaptively discarded, while the rest are retained. 
However, all instances within the bag belong to homologous tissues sampled from the same patient, suggesting potential correlations among these instances.
Therefore, we maximize the following mutual information to improve intra-bag discrimination:
{ \small \begin{equation}
	\label{kkk1}
	\sum_{n\in[N]} \mathbb{I}(\mathbf{B}_n,\overline{\mathbf{B}}_n)=\sum_{n\in[N]}p(\mathbf{B}_n,\overline{\mathbf{B}}_n)\log\frac{p(\mathbf{B}_n|\overline{\mathbf{B}}_n)}{p(\mathbf{B}_n)},
\end{equation}}
where $\overline{\mathbf{B}}_n$ is obtained via the {instance} aggregation network $\mathbb{K}_2$ with the same architecture as $\mathbb{K}_1$, to merge the representations of those discarded instances as $\overline{\mathbf{E}}_n$ that has the definition of $\overline{\mathbf{E}}_n = \mathbb{H}(\mathbf{E}_n,\mathbf{1}-\widehat{\mathbf{h}}_n)$.
To this end, we propose to jointly maximize Eqs.(\ref{kkk2})-(\ref{kkk1}), i.e., 
$max(\sum_{n\in[N]} \mathbb{I}(\mathbf{B}_n,\widetilde{\mathbf{B}}_n)+\mathbb{I}(\mathbf{B}_n, \widehat{\mathbf{B}}_n)+\mathbb{I}(\mathbf{B}_n,\overline{\mathbf B}_n))$, which is formulated by
{ \small
\begin{align}
\label{MI}
	max&(\sum_{n\in[N]}\log\frac{p(\mathbf{B}_n|\widetilde{\mathbf{B}}_n)}{p(\mathbf{B}_n)}+\log\frac{p(\mathbf{B}_n| \widehat{\mathbf{B}}_n)}{p(\mathbf{B}_n)}+\log\frac{p(\mathbf{B}_n|\overline{\mathbf{B}}_n)}{p(\mathbf{B}_n)})\notag\\
	&= max(\sum_{n\in[N]}\log\frac{{p(\mathbf{B}_n|\widetilde{\mathbf{B}}_n)}{p(\mathbf{B}_n|\widehat{\mathbf{B}}_n)}{p(\mathbf{B}_n|\overline{\mathbf{B}}_n)}}{p(\mathbf{B}_n)p(\mathbf{B}_n)p(\mathbf{B}_n)}).
\end{align}}
Instead of constructing a generative model ${{p(\mathbf{B}_n|\widetilde{\mathbf{B}}_n)}{p(\mathbf{B}_n|\widehat{\mathbf{B}}_n)}{p(\mathbf{B}_n|\overline{\mathbf{B}}_n)}}$, we leverage a log-bilinear function $\mathbb{L}$ to model the density ratio which preserves the mutual information:
{ \small
\begin{align}
	\mathbb{L}(\mathbf{B}_n, \widetilde{\mathbf{B}}_n,\widehat{\mathbf{B}}_n, \overline{\mathbf{B}}_n)&=exp(\mathbf{B}_n^T \mathbf W_\mathbb{L} \widetilde{\mathbf{B}}_n)exp(\mathbf{B}_n^T\mathbf W_\mathbb{L} \widehat{\mathbf{B}}_n)exp(\mathbf{B}_n^T \mathbf V_\mathbb{L} \overline{\mathbf{B}}_n)\notag\\
	&=exp(\mathbf{B}_n^T \mathbf W_\mathbb{L} \widetilde{\mathbf{B}}_n+\mathbf{B}_n^T\mathbf W_\mathbb{L}\widehat{\mathbf{B}}_n+\mathbf{B}_n^T \mathbf V_\mathbb{L} \overline{\mathbf{B}}_n)\notag\\
	&\propto{\frac{{p(\mathbf{B}_n|\widetilde{\mathbf{B}}_n)}{p(\mathbf{B}_n|\widehat{\mathbf{B}}_n)}{p(\mathbf{B}_n|\overline{\mathbf{B}}_n)}}{p(\mathbf{B}_n)p(\mathbf{B}_n)p(\mathbf{B}_n)}},
\end{align}}
where $\mathbf W_\mathbb{L}$ and $\mathbf V_\mathbb{L}$ are trainable parameters.
Motivated by \cite{chen2020simple}, we design the triple-tier contrastive learning loss based on the InfoNCE function, which has the following definition:
{ \small \begin{equation}
	\label{tcl}
	\mathcal{L}_{tcl}=-\underset{\mathcal{B}}{\boldsymbol{E}}\left[\log\frac{\mathbb{L}(\mathbf{B}_n,\widetilde{\mathbf{B}}_n,\widehat{\mathbf{B}}_n,{\overline{\mathbf{B}}_n})}{{{\mathbb{L}(\mathbf{B}_n,\widetilde{\mathbf{B}}_n,\widehat{\mathbf{B}}_n,\overline{\mathbf{B}}_n)}+\sum_{i\in{[N]},i\neq n}\mathbb{L}(\mathbf{B}_i,\underline{\mathbf{B}}, \widehat{\mathbf{B}}_n,\overline{\mathbf{B}}_n)}}\right],
\end{equation}}
where $\mathcal{B}$ denotes the collection of all bags.
We provide proof that decreasing $\mathcal{L}_{tcl}$ is equivalent to increasing the lower bound on the Eq.(\ref{MI})(in \textit{Supplementary Materials}).
Furthermore, we introduce the following absolute distance constraint to make bag representation more compact:
{ \small \begin{equation}
    \begin{aligned}
    \mathcal{L}_{adc}=&\sum\limits_{i\in[N]}\max(0,\operatorname{Cos}({{\mathbf{B}}_{n}},\mathbf{\underline{B}})-\operatorname{Cos}({{\mathbf{B}}_{n}},\mathbf{\widetilde{B}_{n}})+\kappa)\\&+{1-\operatorname{Cos}({{\mathbf{B}}_{n}},{{\widehat{\mathbf{B}}}_{n}})}+1-\operatorname{Cos}({{\mathbf{B}}_{n}},{{\overline{\mathbf{B}}}_{n}}),
    \end{aligned}
\end{equation}}
where $\kappa = 1$ refers to the margin that indicates the distance between the distribution of tumor tissue and normal tissue.

We utilize the negative log-partial likelihood of Cox proportional hazard regression model to train the network, which is computed by
{ \small \begin{equation}
    \begin{aligned}
    \mathcal{L}_{cox} &=-\log(\prod\limits_{n:{{\boldsymbol \delta}_{n}}=1}{\frac{exp({\mathbf{B}_n\mathbf{W}_\mathbb{S}})}{\sum\nolimits_{j \in R(\mathbf T_n)}{{exp({{\mathbf{B}}_j\mathbf{W}_\mathbb{S}})}}}})
    \\&=-\sum\limits_{n:{{\boldsymbol \delta}_{n}}=1}{(\mathbf{B}_n\mathbf{W}_\mathbb{S}-\log(\sum\nolimits_{j \in {R}(\mathbf T_n)}{{exp({{\mathbf{B}}_j\mathbf{W}_\mathbb{S}})}}))},
    \end{aligned}
\end{equation}}
where $\mathbf{W}_\mathbb{S}$ is the linear projection matrix in $\mathbb{S}$, and ${R(\mathbf T_n)}=\{i:\mathbf{T}_i\geq \mathbf{T}_n\}$ is the collection of all patients (including censored and uncensored ones) with survival time being longer than $\mathbf T_n$.
Finally, the loss function $\mathcal L_{\textrm{II}}$ for the second curriculum can be defined as
{ \small \begin{equation}
	\mathcal L_{\textrm{II}}=\mathcal{L}_{cox}+\beta_{tcl}\mathcal{L}_{tcl}{ + \beta_{adc} \mathcal{L}_{adc}} + \beta_{s}\mathcal{L}_{s},
\end{equation}}
where $ \beta_{tcl}$, $\beta_{adc}$ and $\beta_{s}$ are the weight coefficients.

\section{Experiments}
In this section, we first describe the data and code availability to ensure reproducibility. Subsequently, we perform a comprehensive performance evaluation and comparative analysis against multiple state-of-the-art models. To further verify the contribution of each module, we conduct ablation experiments under consistent experimental settings.
Beyond quantitative evaluation, we further explore the interpretability of DCMIL by visualizing saliency maps across multi-magnification WSIs and analyzing representative soft-bag instance embeddings.

\subsection{Performance evaluation and comparative analysis}
\begingroup
\newcommand{\cindexcell}[3]{\shortstack{#1{\footnotesize$\pm$#2}\\{\scriptsize #3}}}
\newcommand{\bestcindexcell}[3]{\shortstack{\textbf{#1}{\footnotesize$\pm$#2}\\{\scriptsize #3}}}

\begin{table}
    \caption{Performance comparison across 12 cancer cohorts. Each cell reports the mean $\pm$ std and bootstrap 95\% confidence interval on separate lines. {The $p$-values compare DCMIL with each baseline using paired $t$-tests on fold-level C-indices (12 cohorts $\times$ 5 folds), matched by cohort and fold.}}
    \label{tab:results}
    \centering
    {\renewcommand{\arraystretch}{2}
    \resizebox{\linewidth}{!}{%
    \begin{tabular}{|c|c|*{12}{c|}c|}
\hline
\textbf{Category} & \textbf{Method} & \textbf{KIRC} & \textbf{LUAD} & \textbf{LUSC} & \textbf{STAD} & \textbf{THCA} & \textbf{HNSC} & \textbf{LIHC} & \textbf{BRCA} & \textbf{BLCA} & \textbf{COAD} & \textbf{OV} & \textbf{UCEC} & \textbf{$p$-value} \\
\hline
\multirow{4}{*}{\shortstack{Weakly\\Supervised\\Learning}}
& AMISL
& \cindexcell{0.659}{.021}{[0.641, 0.674]}
& \cindexcell{0.624}{.025}{[0.606, 0.642]}
& \cindexcell{0.614}{.035}{[0.587, 0.642]}
& \cindexcell{0.625}{.044}{[0.590, 0.658]}
& \cindexcell{0.812}{.093}{[0.740, 0.884]}
& \cindexcell{0.624}{.053}{[0.582, 0.661]}
& \cindexcell{0.653}{.033}{[0.627, 0.679]}
& \cindexcell{0.650}{.043}{[0.616, 0.683]}
& \cindexcell{0.625}{.044}{[0.593, 0.663]}
& \cindexcell{0.639}{.094}{[0.559, 0.708]}
& \cindexcell{0.591}{.027}{[0.570, 0.612]}
& \cindexcell{0.688}{.068}{[0.629, 0.736]}
& 4.938E-05 \\
& PGCN
& \cindexcell{0.565}{.089}{[0.496, 0.631]}
& \cindexcell{0.532}{.043}{[0.504, 0.571]}
& \cindexcell{0.578}{.048}{[0.535, 0.614]}
& \cindexcell{0.590}{.060}{[0.543, 0.638]}
& \cindexcell{0.809}{.084}{[0.745, 0.877]}
& \cindexcell{0.605}{.064}{[0.556, 0.649]}
& \cindexcell{0.605}{.071}{[0.555, 0.668]}
& \cindexcell{0.599}{.040}{[0.564, 0.626]}
& \cindexcell{0.594}{.030}{[0.571, 0.617]}
& \cindexcell{0.639}{.047}{[0.602, 0.674]}
& \cindexcell{0.540}{.030}{[0.517, 0.563]}
& \cindexcell{0.613}{.077}{[0.549, 0.675]}
& 1.273E-07 \\
& MesoNet
& \cindexcell{0.643}{.022}{[0.626, 0.661]}
& \cindexcell{0.627}{.020}{[0.610, 0.641]}
& \cindexcell{0.610}{.032}{[0.587, 0.636]}
& \cindexcell{0.648}{.024}{[0.630, 0.670]}
& \cindexcell{0.837}{.023}{[0.819, 0.854]}
& \cindexcell{0.624}{.062}{[0.577, 0.671]}
& \cindexcell{0.676}{.015}{[0.664, 0.688]}
& \cindexcell{0.688}{.015}{[0.676, 0.699]}
& \cindexcell{0.643}{.045}{[0.604, 0.670]}
& \cindexcell{0.689}{.014}{[0.677, 0.700]}
& \cindexcell{0.603}{.017}{[0.590, 0.616]}
& \cindexcell{0.697}{.048}{[0.654, 0.727]}
& 4.938E-05 \\
& HIPT
& \cindexcell{0.646}{.034}{[0.619, 0.673]}
& \cindexcell{0.593}{.037}{[0.560, 0.619]}
& \cindexcell{0.591}{.064}{[0.542, 0.642]}
& \cindexcell{0.650}{.016}{[0.637, 0.662]}
& \cindexcell{0.864}{.079}{[0.801, 0.926]}
& \cindexcell{0.599}{.068}{[0.547, 0.652]}
& \cindexcell{0.621}{.046}{[0.586, 0.656]}
& \cindexcell{0.652}{.043}{[0.622, 0.685]}
& \cindexcell{0.611}{.014}{[0.601, 0.622]}
& \cindexcell{0.630}{.059}{[0.581, 0.672]}
& \cindexcell{0.594}{.020}{[0.580, 0.611]}
& \cindexcell{0.671}{.048}{[0.633, 0.706]}
& 1.013E-04 \\
\hline
\multirow{5}{*}{\shortstack{Whole\\Slide\\Foundation}}
& TITAN
& \cindexcell{0.667}{.032}{[0.642, 0.692]}
& \cindexcell{0.592}{.026}{[0.576, 0.615]}
& \cindexcell{0.505}{.008}{[0.499, 0.511]}
& \cindexcell{0.635}{.022}{[0.618, 0.652]}
& \cindexcell{0.732}{.076}{[0.672, 0.792]}
& \cindexcell{0.593}{.025}{[0.575, 0.611]}
& \cindexcell{0.680}{.023}{[0.662, 0.699]}
& \cindexcell{0.631}{.014}{[0.620, 0.642]}
& \cindexcell{0.628}{.022}{[0.611, 0.645]}
& \cindexcell{0.713}{.026}{[0.693, 0.732]}
& \cindexcell{0.588}{.014}{[0.578, 0.599]}
& \cindexcell{0.707}{.031}{[0.683, 0.731]}
& 7.253E-03 \\
& GigaPath
& \cindexcell{0.675}{.032}{[0.648, 0.699]}
& \cindexcell{0.537}{.008}{[0.530, 0.542]}
& \cindexcell{0.554}{.015}{[0.542, 0.566]}
& \cindexcell{0.619}{.013}{[0.609, 0.630]}
& \cindexcell{0.858}{.038}{[0.828, 0.887]}
& \cindexcell{0.631}{.003}{[0.629, 0.633]}
& \cindexcell{0.643}{.039}{[0.613, 0.673]}
& \cindexcell{0.612}{.007}{[0.606, 0.617]}
& \cindexcell{0.643}{.028}{[0.622, 0.665]}
& \cindexcell{0.694}{.030}{[0.671, 0.719]}
& \cindexcell{0.571}{.014}{[0.561, 0.583]}
& \cindexcell{0.708}{.026}{[0.685, 0.724]}
& 3.411E-03 \\
& CHIEF
& \cindexcell{0.672}{.029}{[0.650, 0.694]}
& \cindexcell{0.538}{.008}{[0.531, 0.543]}
& \cindexcell{0.566}{.005}{[0.562, 0.570]}
& \cindexcell{0.621}{.020}{[0.606, 0.639]}
& \cindexcell{0.795}{.066}{[0.746, 0.844]}
& \cindexcell{0.609}{.015}{[0.597, 0.621]}
& \cindexcell{0.632}{.045}{[0.599, 0.666]}
& \cindexcell{0.618}{.030}{[0.598, 0.644]}
& \cindexcell{0.629}{.020}{[0.617, 0.647]}
& \cindexcell{0.651}{.033}{[0.630, 0.680]}
& \cindexcell{0.570}{.024}{[0.550, 0.587]}
& \cindexcell{0.711}{.025}{[0.692, 0.731]}
& 1.995E-04 \\
& PRISM
& \cindexcell{0.625}{.032}{[0.600, 0.650]}
& \cindexcell{0.545}{.014}{[0.534, 0.556]}
& \cindexcell{0.558}{.032}{[0.535, 0.584]}
& \cindexcell{0.591}{.009}{[0.586, 0.599]}
& \cindexcell{0.804}{.067}{[0.755, 0.856]}
& \cindexcell{0.575}{.016}{[0.563, 0.589]}
& \cindexcell{0.671}{.020}{[0.657, 0.688]}
& \cindexcell{0.617}{.027}{[0.597, 0.640]}
& \cindexcell{0.634}{.024}{[0.612, 0.648]}
& \cindexcell{0.600}{.026}{[0.581, 0.621]}
& \cindexcell{0.598}{.018}{[0.586, 0.613]}
& \cindexcell{0.648}{.033}{[0.622, 0.674]}
& 1.168E-04 \\
& MADELEINE
& \cindexcell{0.657}{.023}{[0.639, 0.675]}
& \cindexcell{0.575}{.029}{[0.553, 0.597]}
& \cindexcell{0.545}{.017}{[0.533, 0.560]}
& \cindexcell{0.633}{.018}{[0.617, 0.646]}
& \cindexcell{0.687}{.053}{[0.647, 0.726]}
& \cindexcell{0.613}{.016}{[0.599, 0.623]}
& \cindexcell{0.676}{.018}{[0.662, 0.690]}
& \cindexcell{0.519}{.009}{[0.512, 0.526]}
& \cindexcell{0.628}{.035}{[0.600, 0.654]}
& \cindexcell{0.702}{.014}{[0.691, 0.712]}
& \cindexcell{0.575}{.012}{[0.567, 0.585]}
& \cindexcell{0.703}{.027}{[0.681, 0.724]}
& 7.253E-03 \\
\hline
\multirow{1}{*}{Proposed}
& DCMIL
& \bestcindexcell{0.676}{.018}{[0.662, 0.690]}
& \bestcindexcell{0.646}{.020}{[0.631, 0.661]}
& \bestcindexcell{0.659}{.024}{[0.640, 0.677]}
& \bestcindexcell{0.672}{.017}{[0.661, 0.686]}
& \bestcindexcell{0.871}{.052}{[0.838, 0.918]}
& \bestcindexcell{0.667}{.016}{[0.655, 0.680]}
& \bestcindexcell{0.681}{.022}{[0.664, 0.697]}
& \bestcindexcell{0.715}{.017}{[0.702, 0.729]}
& \bestcindexcell{0.669}{.047}{[0.637, 0.710]}
& \bestcindexcell{0.718}{.030}{[0.702, 0.746]}
& \bestcindexcell{0.616}{.022}{[0.600, 0.636]}
& \bestcindexcell{0.726}{.035}{[0.698, 0.753]}
& -- \\
\hline
\end{tabular}%
    }}
    \vspace{-10pt}
\end{table}

DCMIL enables direct transformation of gigapixel-size WSIs into outcome predictions, bypassing the need for manual region selection or patch-level annotations.
We evaluated DCMIL on 12 cancer types using a 5-fold cross-validation strategy across 5,709 patients and approximately 12.54 million image tiles.
Specifically, we divided the entire dataset into five folds, among which four folds for model training and the remaining one for model evaluation. 
Within the training set, we further split out one-fifth for validation to select the model's hyperparameters. 
Notably, the cross-validation strategy was conducted on the patient level to prevent data leakage, which means that the WSI of each patient only appeared in one of these subsets.
To further quantify the uncertainty of prognostic performance, we calculated 95\% confidence intervals for the C-index using 2,000 patient-level bootstrap replicates within each cancer cohort. Patients, together with their survival outcomes and corresponding model predictions, were resampled with replacement, and the C-index was recalculated for each bootstrap sample. The resulting bootstrap distributions were used to derive the 95\% confidence intervals reported in Table~\ref{tab:results}. {
Using paired $t$-tests~\cite{ross2017paired}, the $p$-values were calculated from fold-level C-index values pooled across 12 cancer cohorts (five folds per cohort), comparing DCMIL with each baseline method and pairing by cohort and fold.
}

Prognostic performance was assessed using the concordance index (C-index) and Kaplan–Meier (KM) survival analysis. 
DCMIL was compared against conventional weakly supervised learning methods, including graph-based PGCN~\cite{chen2021whole}, transformer-based HIPT~\cite{chen2022scaling}, and multi-instance learning models such as MesoNet~\cite{courtiol2019deep} and AMISL~\cite{yao2020whole}. We additionally included recent whole-slide foundation models, including TITAN~\cite{ding2025multimodal}, GigaPath~\cite{xu2024whole}, CHIEF~\cite{wang2024pathology}, PRISM~\cite{vorontsov2026end}, and MADELEINE~\cite{jaume2024multistain}, as contextual benchmarks.
It should be noted that these two groups of methods follow different representation-learning paradigms. The foundation models rely on large-scale external pretraining to obtain general-purpose pathology representations, which are subsequently adapted to cancer-specific survival prediction on the target cohorts. In contrast, DCMIL does not rely on large-scale pan-cancer pretraining; instead, its instance representations are learned directly from the training portion of each target cohort using prognosis-oriented weak supervision in Curriculum I, followed by patient-level survival modeling in Curriculum II. Therefore, the results in Table~\ref{tab:results} should be interpreted as a comparison of task-level prognostic performance under the adopted downstream evaluation protocol, rather than as a controlled comparison of intrinsic representation quality or a direct demonstration of superiority over pathology foundation models.

Under this evaluation setting, DCMIL achieved competitive performance across the 12 cancer cohorts, with a mean C-index of 0.693. Patient-level bootstrap analysis with 2,000 resamples further supported the stability of the performance estimates. For DCMIL, the 95\% confidence intervals of the C-index were relatively narrow across most cancer cohorts, with interval widths below 0.05 in 9 of the 12 datasets. Although the bootstrap intervals overlapped with those of the strongest competing methods in several cohorts, DCMIL maintained consistently competitive C-index estimates across the 12 cancer types.
{The Kaplan--Meier curves and corresponding $p$-values (Figure~\ref{fig_km}) further illustrate the survival separation between the predicted high- and low-risk groups.}
\begin{figure}[t]
	\centering
	\includegraphics[width=0.8\textwidth]{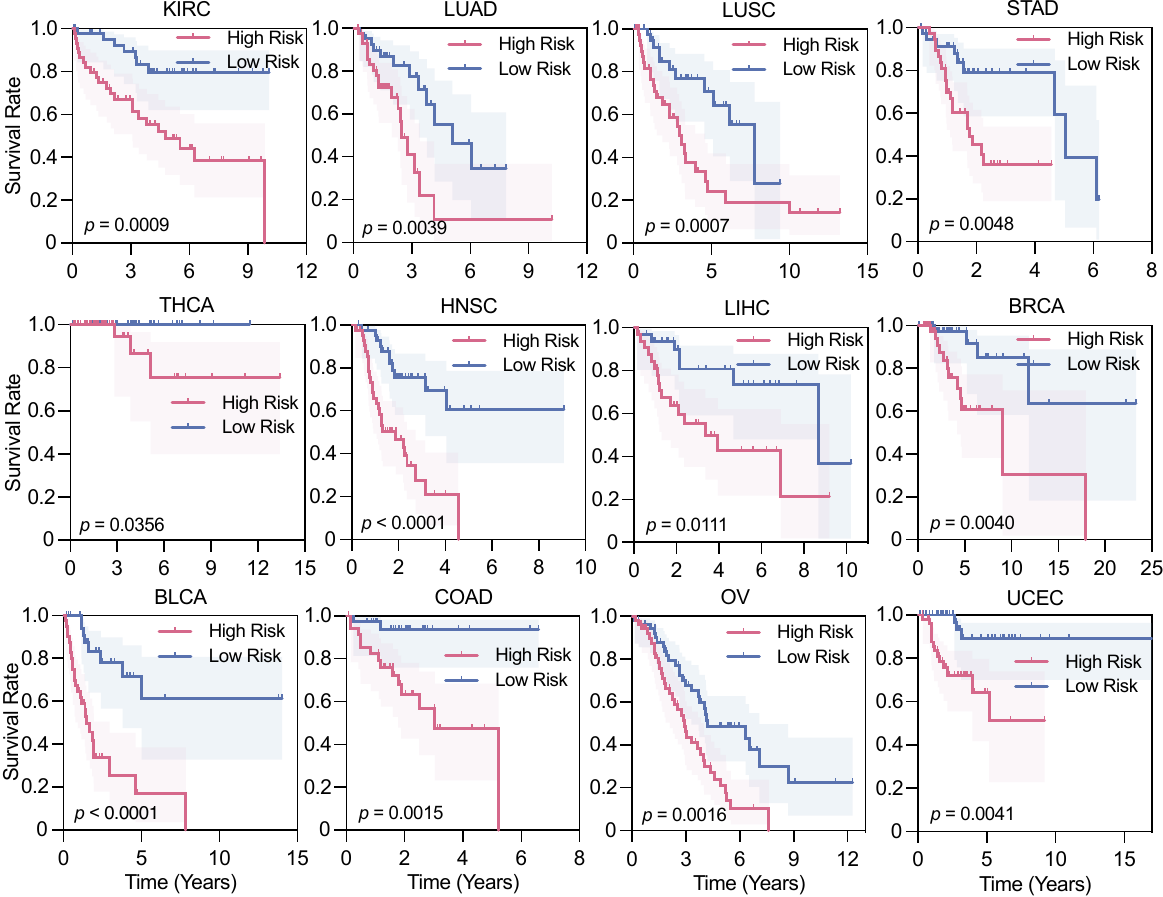}
	\caption{\textbf{The KM curves along with p-values on twelve datasets.}}
	\label{fig_km}
    \vspace{-10pt}
\end{figure}

\subsection{Ablation experiments}
{
We validated the efficacy of crucial components in Curriculum I (C-I) and Curriculum II (C-II) through systematic ablation experiments. In C-I, we evaluated the contributions of multi-magnification learning (MM), saliency-guided learning (SG), hierarchical transfer (HT), and pairwise ranking loss (PR). We additionally introduced a self-supervised C-I variant (SSL-C-I), in which the survival-derived weak supervision was replaced with semantically relevant contrastive learning (SRCL) without survival-derived labels. In C-II, we evaluated soft-bag learning (SB), constrained self-attention (CSA), triple-tier contrastive learning (TCL), and the absolute distance constraint (ADC).

As shown in Table~\ref{tab_abl}, the C-II-only baseline achieved an average C-index of 0.666, while progressively incorporating the C-I components improved overall performance. Removing SG, HT, or PR reduced the average C-index to 0.676, 0.680, and 0.679, respectively. SSL-C-I achieved a competitive average C-index of 0.678, compared with 0.693 for the full model, indicating that self-supervised morphological representations remain informative while survival-derived weak supervision provides additional prognosis-relevant guidance. For C-II, the C-I-only configuration achieved 0.675; incorporating SB and CSA increased the performance, while removing TCL together with ADC reduced the average C-index to 0.675. Removing ADC alone decreased the average C-index from 0.693 to 0.689, further supporting the complementary contributions of the C-II components.}


\begin{figure}[ht!]
    \centering
    \includegraphics[width=0.8\textwidth]{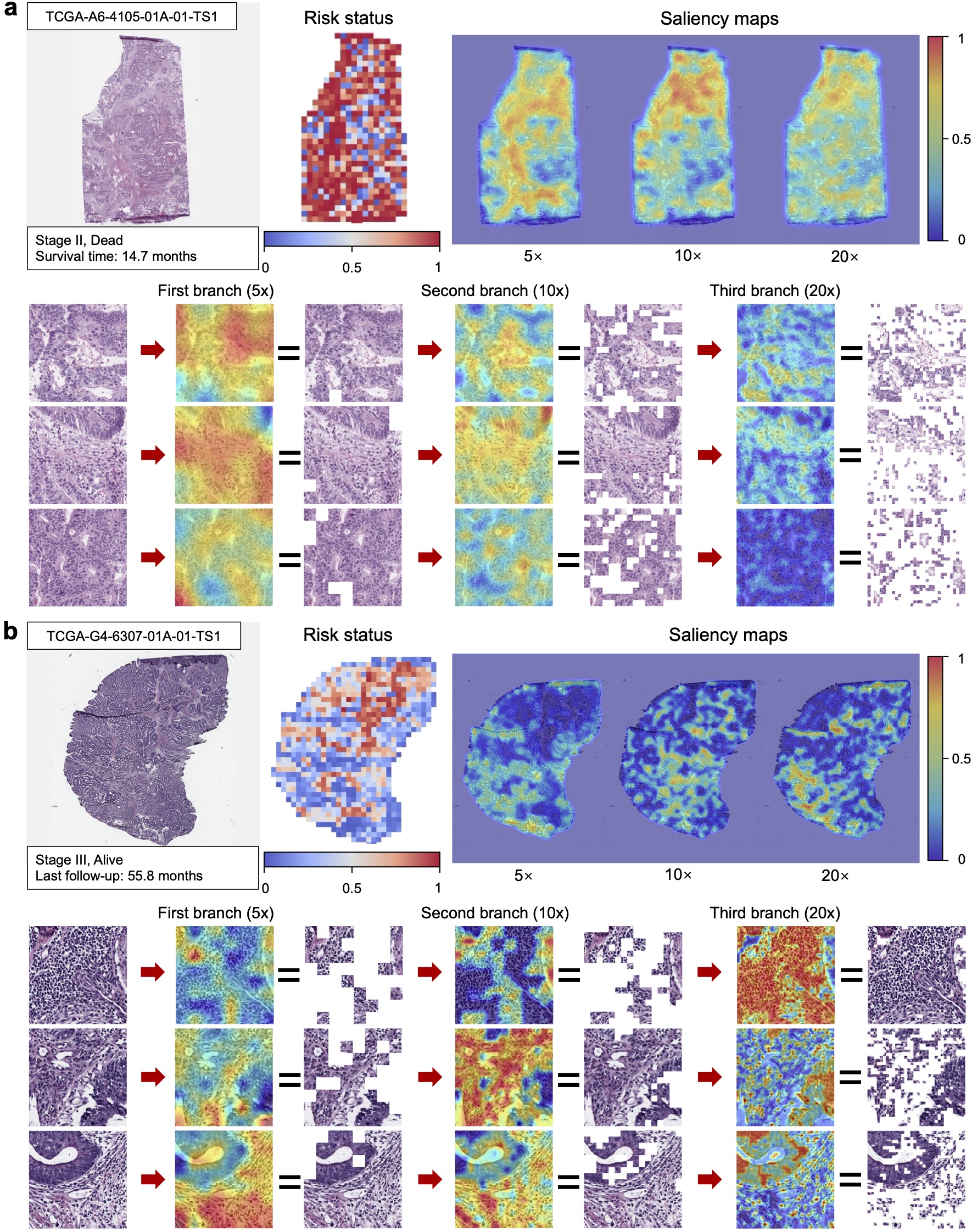}
    \caption{\textbf{Visualizations of saliency maps across multi-magnification WSIs on Curriculum I for COAD dataset.} Comparisons between high-risk (\textbf{a}) and low-risk (\textbf{b}) cases. \textbf{Row 1}: Original WSI, risk estimation, and saliency at 5$\times$--20$\times$. \textbf{Row 2}: Low-magnification branch outputs for representation guidance.
	}
    \vspace{-10pt}
    \label{fig_2}
\end{figure}

\begin{table}[t]
\footnotesize
\caption{\textbf{The results of ablation experiments on twelve datasets}. 
The boldface denotes the best result.}
\label{tab_abl}
\centering
\setlength{\tabcolsep}{0.7mm}
\begin{threeparttable}
\resizebox{\linewidth}{!}{
\begin{tabular}{ccccc|cccccccccccc}
\toprule
MM & SG & HT & PR & \multicolumn{1}{|c|}{C-II} &
KIRC & LUAD & LUSC & STAD & THCA & HNSC & LIHC & BRCA & BLCA & COAD & OV & UCEC \\
\midrule

\XSolidBrush & \XSolidBrush & \XSolidBrush & \XSolidBrush & \Checkmark &
0.653\footnotesize$\pm$.013 &
0.632\footnotesize$\pm$.016 &
0.629\footnotesize$\pm$.016 &
0.649\footnotesize$\pm$.017 &
0.831\footnotesize$\pm$.049 &
0.647\footnotesize$\pm$.035 &
0.649\footnotesize$\pm$.046 &
0.678\footnotesize$\pm$.030 &
0.653\footnotesize$\pm$.024 &
0.684\footnotesize$\pm$.036 &
0.597\footnotesize$\pm$.014 &
0.686\footnotesize$\pm$.036 \\

\Checkmark & \XSolidBrush & \Checkmark & \Checkmark & \Checkmark &
0.658\footnotesize$\pm$.022 &
0.635\footnotesize$\pm$.008 &
0.634\footnotesize$\pm$.014 &
0.663\footnotesize$\pm$.027 &
0.841\footnotesize$\pm$.041 &
0.649\footnotesize$\pm$.043 &
0.668\footnotesize$\pm$.038 &
0.698\footnotesize$\pm$.018 &
0.659\footnotesize$\pm$.017 &
0.707\footnotesize$\pm$.044 &
0.604\footnotesize$\pm$.025 &
0.691\footnotesize$\pm$.022 \\

\Checkmark & \Checkmark & \XSolidBrush & \Checkmark & \Checkmark &
0.667\footnotesize$\pm$.035 &
0.642\footnotesize$\pm$.033 &
0.642\footnotesize$\pm$.023 &
0.648\footnotesize$\pm$.019 &
0.846\footnotesize$\pm$.031 &
0.660\footnotesize$\pm$.035 &
0.678\footnotesize$\pm$.035 &
0.689\footnotesize$\pm$.034 &
0.663\footnotesize$\pm$.033 &
0.709\footnotesize$\pm$.031 &
0.615\footnotesize$\pm$.025 &
0.698\footnotesize$\pm$.063 \\

\Checkmark & \Checkmark & \Checkmark & \XSolidBrush & \Checkmark &
0.661\footnotesize$\pm$.026 &
0.644\footnotesize$\pm$.025 &
0.636\footnotesize$\pm$.023 &
0.650\footnotesize$\pm$.023 &
0.858\footnotesize$\pm$.061 &
0.653\footnotesize$\pm$.041 &
0.678\footnotesize$\pm$.041 &
0.671\footnotesize$\pm$.017 &
0.665\footnotesize$\pm$.019 &
0.716\footnotesize$\pm$.041 &
0.612\footnotesize$\pm$.022 &
0.698\footnotesize$\pm$.034 \\

\midrule
\multicolumn{5}{c|}{SSL-C-I + C-II} &
0.658\footnotesize$\pm$.013 &
0.632\footnotesize$\pm$.016 &
0.646\footnotesize$\pm$.021 &
0.649\footnotesize$\pm$.021 &
0.859\footnotesize$\pm$.039 &
0.655\footnotesize$\pm$.036 &
0.667\footnotesize$\pm$.041 &
0.707\footnotesize$\pm$.024 &
0.647\footnotesize$\pm$.021 &
0.697\footnotesize$\pm$.046 &
0.613\footnotesize$\pm$.014 &
0.709\footnotesize$\pm$.027 \\

\midrule\midrule
\multicolumn{1}{c|}{C-I} & SB & CSA & TCL & ADC \\
\midrule

\Checkmark & \XSolidBrush & \XSolidBrush & \XSolidBrush & \XSolidBrush &
0.667\footnotesize$\pm$.037 &
0.634\footnotesize$\pm$.016 &
0.649\footnotesize$\pm$.021 &
0.648\footnotesize$\pm$.043 &
0.833\footnotesize$\pm$.044 &
0.647\footnotesize$\pm$.016 &
0.659\footnotesize$\pm$.036 &
0.698\footnotesize$\pm$.051 &
0.654\footnotesize$\pm$.033 &
0.691\footnotesize$\pm$.033 &
0.610\footnotesize$\pm$.027 &
0.711\footnotesize$\pm$.080 \\

\Checkmark & \Checkmark & \XSolidBrush & \Checkmark & \Checkmark &
0.673\footnotesize$\pm$.025 &
0.646\footnotesize$\pm$.010 &
0.656\footnotesize$\pm$.016 &
0.661\footnotesize$\pm$.015 &
0.862\footnotesize$\pm$.035 &
0.654\footnotesize$\pm$.015 &
0.676\footnotesize$\pm$.033 &
0.699\footnotesize$\pm$.024 &
0.665\footnotesize$\pm$.011 &
0.712\footnotesize$\pm$.046 &
0.608\footnotesize$\pm$.019 &
0.714\footnotesize$\pm$.026 \\

\Checkmark & \Checkmark & \Checkmark & \XSolidBrush & \XSolidBrush &
0.663\footnotesize$\pm$.042 &
0.637\footnotesize$\pm$.004 &
0.640\footnotesize$\pm$.026 &
0.652\footnotesize$\pm$.021 &
0.846\footnotesize$\pm$.070 &
0.651\footnotesize$\pm$.018 &
0.663\footnotesize$\pm$.044 &
0.671\footnotesize$\pm$.053 &
0.660\footnotesize$\pm$.011 &
0.704\footnotesize$\pm$.036 &
0.606\footnotesize$\pm$.016 &
0.704\footnotesize$\pm$.019 \\

\Checkmark & \Checkmark & \Checkmark & \Checkmark & \XSolidBrush &
0.675\footnotesize$\pm$.036 &
0.640\footnotesize$\pm$.010 &
0.645\footnotesize$\pm$.019 &
0.666\footnotesize$\pm$.022 &
0.869\footnotesize$\pm$.051 &
0.657\footnotesize$\pm$.040 &
0.679\footnotesize$\pm$.047 &
0.713\footnotesize$\pm$.020 &
0.667\footnotesize$\pm$.029 &
0.717\footnotesize$\pm$.027 &
0.614\footnotesize$\pm$.031 &
0.721\footnotesize$\pm$.021 \\

\Checkmark & \Checkmark & \Checkmark & \Checkmark & \Checkmark &
\textbf{0.676}\footnotesize$\pm$.018 &
\textbf{0.646}\footnotesize$\pm$.020 &
\textbf{0.659}\footnotesize$\pm$.024 &
\textbf{0.672}\footnotesize$\pm$.017 &
\textbf{0.871}\footnotesize$\pm$.052 &
\textbf{0.667}\footnotesize$\pm$.016 &
\textbf{0.681}\footnotesize$\pm$.022 &
\textbf{0.715}\footnotesize$\pm$.017 &
\textbf{0.669}\footnotesize$\pm$.047 &
\textbf{0.718}\footnotesize$\pm$.030 &
\textbf{0.616}\footnotesize$\pm$.022 &
\textbf{0.726}\footnotesize$\pm$.035 \\
\bottomrule
\end{tabular}}

\flushleft
\footnotesize
\textit{Note:} C-I, Curriculum I; MM, multi-magnification strategy; SG, saliency-guided method; HT, hierarchical transfer strategy; PR, pairwise ranking loss; C-II, Curriculum II; SB, soft-bag learning; CSA, constrained self-attention module; TCL, triple-tier contrastive learning; ADC, absolute distance constraint. SSL-C-I, a self-supervised C-I variant using semantically relevant contrastive learning (SRCL) without survival-derived labels.
\end{threeparttable}
\vspace{-10pt}
\end{table}

\subsection{Saliency maps across multi-magnification WSIs}
DCMIL can highlight fine-grained prognosis-salient regions on WSIs.
Figure~\ref{fig_2} illustrates two representative cases from the COAD dataset, including a high-risk case (Figure~\ref{fig_2}a, stage II, deceased at 14.7 months) and a low-risk case (Figure~\ref{fig_2}b, stage III, survived over 55.8 months).
In Figure~\ref{fig_2}a/b, the first row shows original WSI, patch-level risk status estimation, and saliency maps of WSIs at different magnifications (i.e., 5$\times$, 10$\times$, and 20$\times$).
The second row exhibits the outputs derived from the low-magnification branches, which guide fine-grained representation learning at higher magnifications.
From Figure~\ref{fig_2}, we can observe that the lower-stage case exhibits a higher risk, indicating that patients with different tumor stages may exhibit distinct prognostic risk profiles.
It also demonstrates that DCMIL can accurately predict risk stratification status, with the high-risk case exhibiting more extensive regions of high-risk probability than the low-risk case.
The saliency maps of WSIs at different magnifications show that each branch highlights distinct yet complementary regions.
From local tiles, we find that different branches emphasize unique spatial information: 
5$\times$ tiles accentuate textural alterations in local microenvironment, 10$\times$ tiles highlight tissue and cell cluster patterns, and 20$\times$ tiles focus on nuclear morphology. 
This may stem from the token size in the ViT-like architecture, covering approximately 8 $\mu m^2$ (single cell) at 20$\times$ tiles, 32 $\mu m^2$ (tissue or cell cluster) at 10$\times$ tiles, and over 100 $\mu m^2$ (local microenvironment) at 5$\times$ tiles. 
From 5$\times$ tiles, we also notice inconsistent salient regions in homogenous tissues.
This inconsistency likely arises from minimal inter-token variability within such tissues, posing a challenge for learning discriminative representations. 
Even so, the model appears to compensate by leveraging complementary cues from the tiles at other magnifications.
Furthermore, we conjecture that two patterns occur in saliency map guidance: fine-grained representations of higher-magnification tiles are learned either from the salient regions at lower magnifications or from expanded regions that compensate for lower-magnification saliency map guidance.
As exhibited in Figure~\ref{fig_2}a (second row), fine-grained representations are derived from smaller salient regions as magnification increases.
As shown in Figure~\ref{fig_2}b (second row), fine-grained representations are extracted from those regions that may not be salient at lower magnifications. 
Actually, such phenomenon occurs as the selection of tiles is revisitable and not entirely determined by the low-magnification saliency map.
This enables flexible learning of fine-grained yet complementary information, while saliency map guidance acts as a dynamic masking strategy to help alleviate overfitting.
\begin{figure}[t!]
    \centering
    \includegraphics[width=0.8\textwidth]{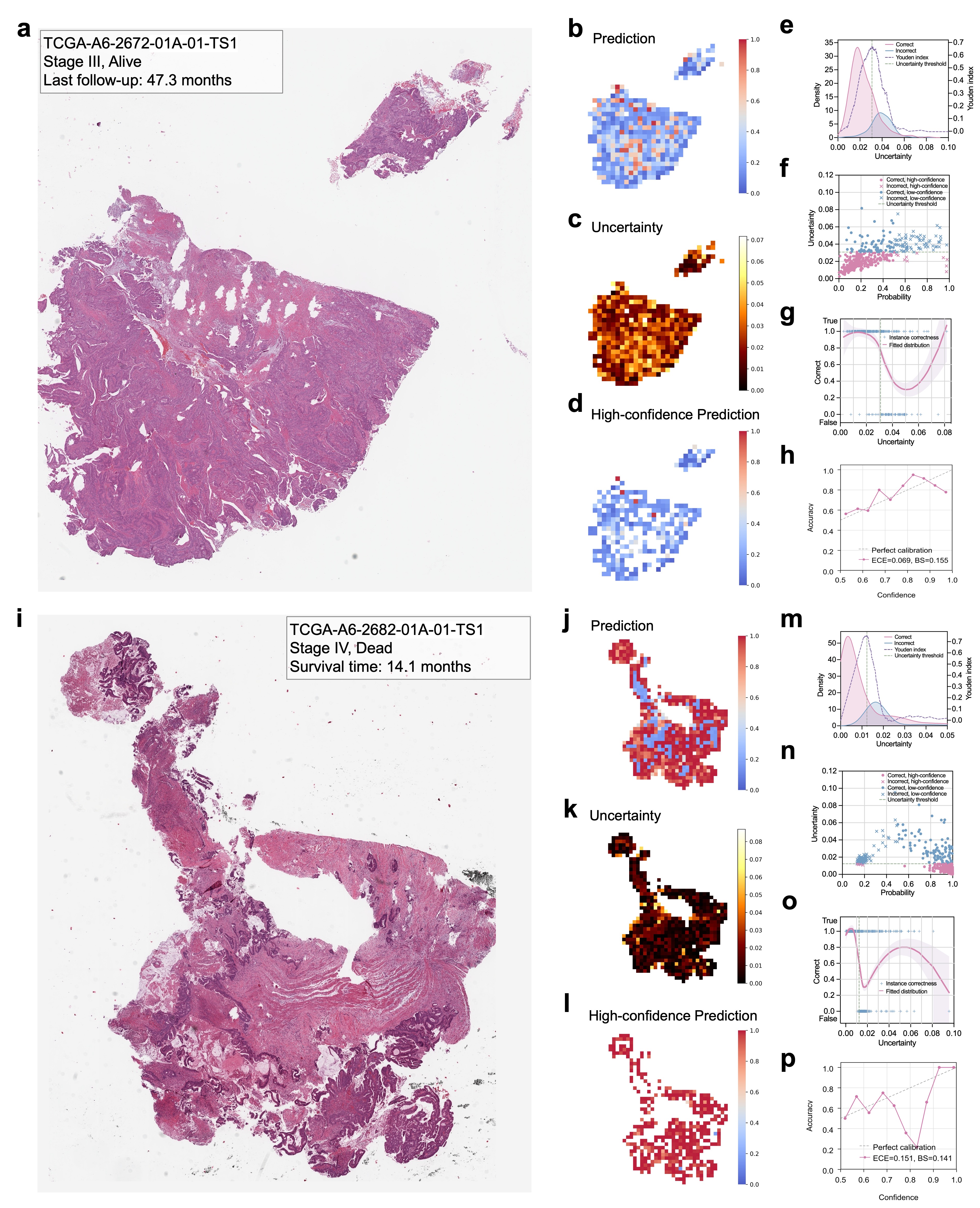}
    \caption{\textbf{Instance uncertainty estimation towards risk stratification on Curriculum I for COAD dataset.}
    This includes a low-risk case (stage III, survived over 47.3 months) and a high-risk case (stage IV, deceased at 14.1 months).
    \textbf{a, i}, Original WSIs. \textbf{b, j}, Risk maps. \textbf{c, k}, Uncertainty maps. \textbf{d, l}, High-confidence instances. \textbf{e, m}, Probability density distributions (Red: correct; Blue: incorrect; Purple: Youden index). \textbf{f, n}, Instance uncertainty distributions. \textbf{g, o}, Correlation between uncertainty and prediction correctness. \textbf{h, p} show the reliability curves.
    }
    \vspace{-10pt}
    \label{fig_4}
\end{figure}
\subsection{Robust instance uncertainty estimation towards risk stratification.}
We performed instance uncertainty estimation towards risk stratification by applying a dropout operator to the network, enabling multiple stochastic forward passes (set to 30) during the prediction stage.
The standard deviation of network output from multiple stochastic forward passes provides an intuitive indication of the uncertainty. 
Given the distribution of instance uncertainty, we determined the uncertainty threshold which optimally separates correct and incorrect predictions by maximizing the Youden index~\cite{dolezal2022uncertainty}. 
Figure~\ref{fig_4} shows two representative cases from the COAD dataset, including a low-risk case (Figure~\ref{fig_4}a-\ref{fig_4}h, stage III, survived over 47.3 months) and a high-risk case (Figure~\ref{fig_4}i-\ref{fig_4}p, stage IV, deceased at 14.1 months). 
Figure~\ref{fig_4}a~and~\ref{fig_4}i show the original WSIs, and Figure~\ref{fig_4}b~and~\ref{fig_4}j present the patch-level risk prediction maps.
The instance uncertainty maps are exhibited in Figure~\ref{fig_4}c~and~\ref{fig_4}k.
We applied the uncertainty threshold to filter out those regions with high uncertainty, masking them on the patch-level risk prediction maps to retain the instances with only high-confidence predictions.
As illustrated in Figure~\ref{fig_4}d~and~\ref{fig_4}l, nearly all instances with high-confidence predictions correspond to those with correct predictions. 
In other words, for low-risk cases, high-confidence predictions cluster in low-risk regions, while for high-risk cases, they cluster in high-risk regions.
Figures~\ref{fig_4}e~and~\ref{fig_4}m display the probability density distributions of correct (red curve) and incorrect (blue curve) predictions, along with Youden index curve (purple, dotted). It is evident that the uncertainty threshold (black dotted line) effectively separates correct from incorrect predictions.
Figures~\ref{fig_4}f~and~\ref{fig_4}n illustrate the distribution of instances with varying levels of uncertainty.
Intuitively, there are fewer high-confidence incorrect predictions (red fork) compared to high-confidence correct predictions (red dot), whereas low-confidence incorrect predictions (blue fork) constitute a large proportion of low-confidence predictions.
Additionally, Figures~\ref{fig_4}g~and~\ref{fig_4}o illustrate the distribution of uncertainty concerning correct and incorrect predictions, which further substantiates that correct predictions are associated with lower uncertainties and vice versa.
Figures~\ref{fig_4}h~and~\ref{fig_4}p show the reliability curves, where the diagonal line denotes perfect calibration, and ECE and BS quantify the calibration performance of instance-level predictions.
{Superior reliability curves (closer to the diagonal) together with lower ECE and Brier Score indicate that the model produces better-calibrated probability estimates, supporting more reliable instance-level risk predictions and improved model interpretability.}
These results suggest that the model has the capability to perform robust instance uncertainty estimation for risk stratification, thereby laying the foundation for Curriculum II.

\subsection{Representative soft-bag instance embedding}
\begin{figure}[t!]
    \centering
    \includegraphics[width=0.8\textwidth]{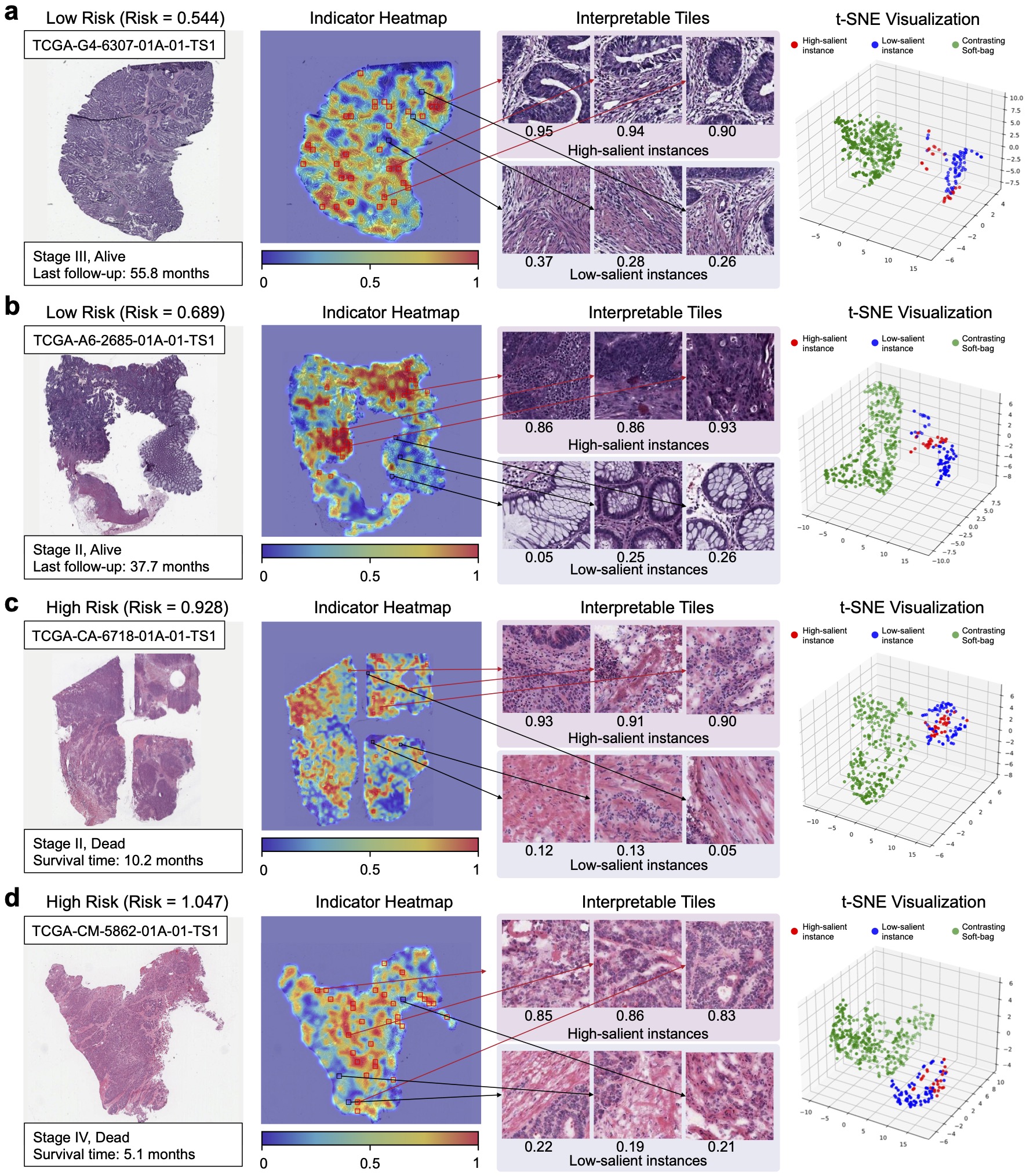}
    \caption{\textbf{Representative COAD cases visualized on Curriculum II.} \textbf{a, b}, Low-risk patients. \textbf{c, d}, High-risk patients. Each panel displays: the original WSI; an indicator heatmap showing soft-bag instance selection and localization; interpretive tiles highlighting high/low-salient instances with scores; and a t-SNE plot depicting embedding distributions of salient instances against contrasting soft-bags.
    }
    \vspace{-10pt}
    \label{fig_3}
\end{figure}
{Human-readable interpretability facilitates examination of whether the model's predictive basis is consistent with pathologists' knowledge and provides a basis for future human-in-the-loop evaluation.}
DCMIL model performs prognosis analysis by first identifying and aggregating representative (i.e., high-salient) regions within the WSI, while contrasting these with other WSIs.
To visualize and interpret representative regions, we generate an indicator heatmap by normalizing the indicator scores and mapping them to their corresponding spatial locations in the WSI.
Figure~\ref{fig_3} presents four cases from the COAD dataset, illustrated with the original WSI, indicator heatmap, interpretable tiles, and t-SNE visualization. 
These cases include two low-risk patients (Figure~\ref{fig_3}a~and~\ref{fig_3}b, stage III and II, survival times over 55.8 and 37.7 months) and two high-risk patients (Figure~\ref{fig_3}c~and~\ref{fig_3}d, stage II and IV, deceased at 10.2 and 5.1 months). 
The indicator heatmap visualizes the soft-bag instance selection and its localization within the WSI, allowing us to observe that these selected representative instances are widely distributed throughout the tumor microenvironment.
Interpretive tiles exhibit several high-salient and low-salient instances, along with their corresponding indicator scores. 
The visual assessment of a board-certified pathologist across various cancer types indicates that the soft-bag captures fine-grained morphological embeddings, often focusing on regions with dense nuclei. 
The t-SNE visualization depicts the embedding distributions of high-salient instances, low-salient instances, and the contrasting soft-bags from other WSIs.
We observe that the selected (high-salient) instances intermingle with the unselected (low-salient) instances within the same WSI and exhibit a consistent distribution, which suggests those high-salient instances can effectively represent the heterogeneous tumor microenvironment of the WSI.
Additionally, the distribution of instances within the same WSI is significantly distinct from that of the contrasting soft-bags.
Such phenomenon underscores the importance of maximizing the mutual information between selected and unselected instances by utilizing selected instances to identify unselected instances from the instance pool of soft-bag.

\begin{figure}[t!]
    \centering
    \includegraphics[width=0.8\textwidth]{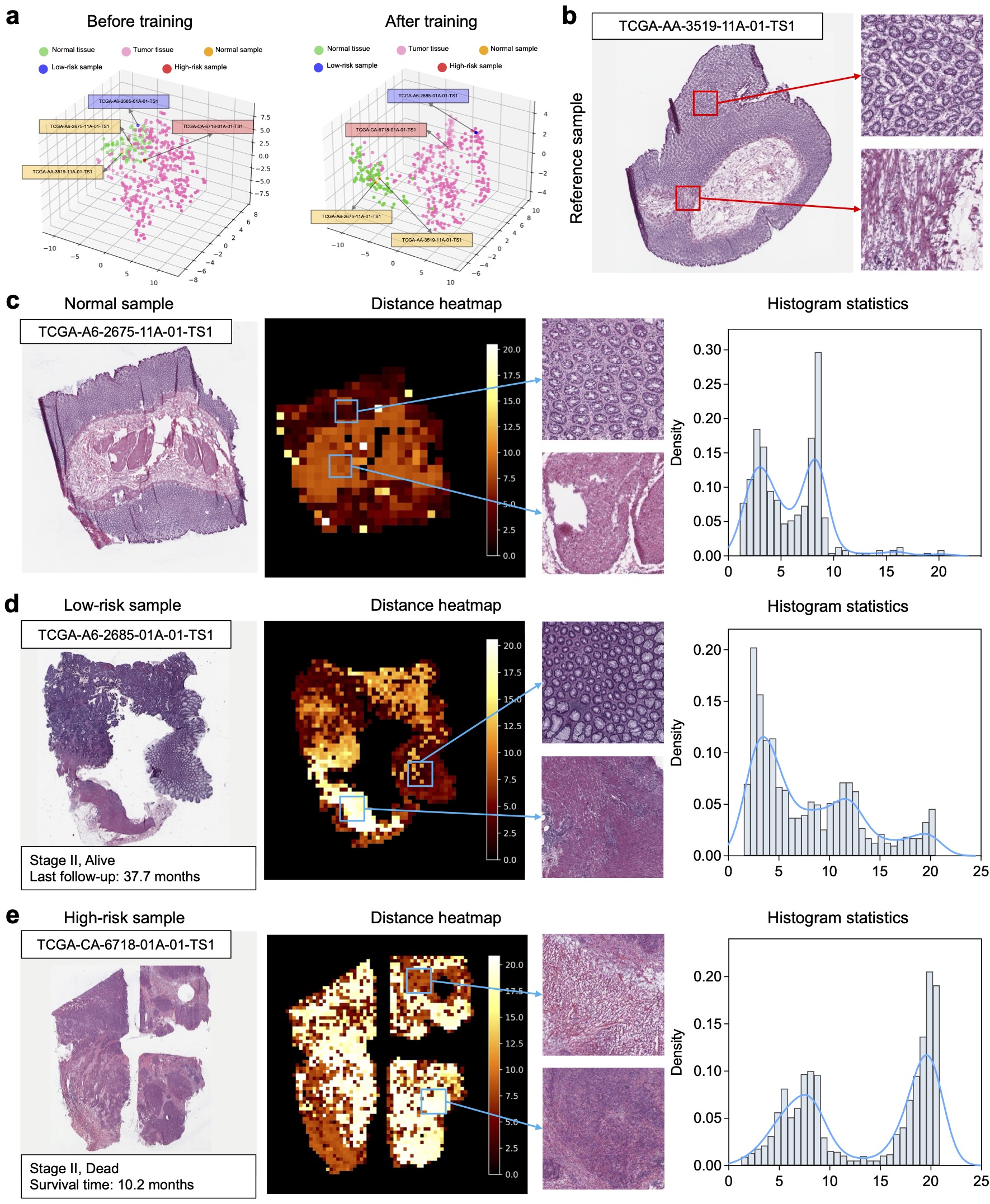}   \caption{\textbf{Visualization of comparative analysis between tumor and normal tissues on Curriculum II for COAD dataset.}
    \textbf{a}, t-SNE distribution plots. \textbf{b}, Normal reference. \textbf{c}--\textbf{e}, Comparisons of normal (\textbf{c}), low-risk (\textbf{d}), and high-risk (\textbf{e}) samples showing original WSIs, distance heatmaps, and histogram statistics.}
    \label{fig_6}
    \vspace{-10pt}
\end{figure}

\subsection{Dissecting tissue heterogeneity between tumor and normal tissues.}
Tumor-adjacent normal tissue may convey significant prognostic information and reveal certain prognostic indicators~\cite{aran2017comprehensive}, including pH levels, stromal behavior, and transcriptomic and epigenetic aberrations.
Comparative analysis against normal tissue facilitates the identification of key morphological malignancy hallmarks, expediting definitive prognosis evaluation~\cite{oh2023transcriptomic}.
To support these advances, we introduced normal tissue slides as control samples for contrastive learning to enhance inter-bag discrimination, as illustrated in Figure~\ref{fig_6}. 
Figure~\ref{fig_6}a presents t-SNE visualization showing the distribution of tumor and normal tissues before and after training. 
We observe that the distribution of normal tissues overlaps with that of tumor tissues before training, and the well-trained DCMIL can effectively distinguish between normal and tumor samples. 
For further analysis, we randomly selected a normal tissue as a reference, as shown in Figure~\ref{fig_6}b. 
Then, we utilized the Euclidean distance to quantify and analyze the similarity/difference between instance-level representations of a normal sample, a low-risk tumor sample, and a high-risk tumor sample compared to the reference one. 
The first column of Figure~\ref{fig_6}c-e exhibits the original WSIs of these samples, the second column illustrates the distance heatmaps along with two representative tiles, and the third column shows the histogram statistics of the distance heatmaps.
The distance heatmaps reveal that the low-risk sample is more dissimilar to the reference sample than the normal sample, but less dissimilar than the high-risk sample.
The histogram statistics show that the distance (or dissimilarity) distribution for the normal sample predominantly falls within 10 units. 
In contrast, for the low-risk sample, the distance is distributed across the entire range, decreasing with increasing distance. For the high-risk sample, the distances primarily cluster within the ranges of 15-20 units and 5-10 units.
Therefore, using normal samples as a reference allows DCMIL to effectively differentiate normal and tumor tissues, and distinguish samples with different risk levels.

\section{DISCUSSION}
This study shows 
{that DCMIL can quantify tumor heterogeneity, enable tractable analysis of gigapixel-size WSIs by directly generating risk-stratification and outcome predictions without dense annotations, achieve competitive prognostic performance across twelve TCGA cancer types, and identify fine-grained prognosis-salient regions. In addition, external evaluation on CPTAC-HNSCC provides evidence of cross-cohort robustness specifically for HNSC.}

In the first curriculum, DCMIL utilizes a low-magnification saliency map to guide the encoding of high-magnification instances, exploring fine-grained information across multi-magnification WSIs, and its quantitative performance improvement is validated in Table~\ref{tab_abl}.
This mimics clinical practice where low-magnification cues guide doctors to significant regions at higher magnifications, supplemented by details observed at higher magnifications. 
The saliency maps (Figure~\ref{fig_2}) and uncertainty maps (Figure~\ref{fig_4}) provide complementary visual evidence for examining prognosis-salient morphological patterns across WSIs and local tiles.
The saliency visualization reveals that multi-scale instance attention focuses on unique spatial survival-related information: 5$\times$ tiles accentuate textural alterations in the local microenvironment, 10$\times$ tiles highlight tissue and cell cluster patterns, and 20$\times$ tiles focus on nuclear morphology, often forming a progressive and complementary relationship.
This approach allows for flexible learning of fine-grained yet complementary information, with saliency map guidance acting as a dynamic masking strategy to reduce overfitting. 

In the second curriculum, unlike general MIL-based approaches that use attention-based pooling and treat different slide locations as independent regions, DCMIL provides the model with the flexibility of selectively aggregating information from WSI, learning potential nonlinear interactions between instances, reducing intra-bag redundancy and enhancing context awareness.
Additionally, it introduces a normal control and employs triple-tier contrastive learning to enhance both intra-bag and inter-bag discrimination, thereby improving the model’s capability in prognosis inference. 
The ablation results in Table~\ref{tab_abl} support the contributions of the Curriculum II components, while the controlled study in Appendix~G further assesses the incremental contribution of tumor--normal contrast.
Indicator heatmaps and t-SNE visualizations (i.e., Figure~\ref{fig_3} on COAD)
underscore the importance of maximizing mutual information between selected and unselected instances by using selected instances to recognize unselected instances from the soft-bag instance pool. 
The comparative analysis between tumor and normal tissues (i.e., Figure~\ref{fig_6}) shows that DCMIL effectively differentiates normal and tumor tissues and distinguishes samples with different risk levels.
Furthermore, the visualization of selected instances provides additional evidence for interpreting the model's aggregation behavior and identifying regions associated with prognostic predictions.

Several limitations remain regarding interpretability and generalizability. Although saliency and uncertainty visualizations highlight prognosis-associated patterns, their biological significance requires further pathological and molecular validation. Model performance may also be influenced by hyperparameter choices and variations in patient populations, tissue preparation, staining protocols, and image acquisition. While independent validation on CPTAC-HNSCC provides evidence of cross-cohort robustness for HNSC, broader external validation across cancer types and clinical environments remains necessary.

\section{Conclusion}


In this study, we present an extended DCMIL framework for weakly supervised cancer prognosis analysis with whole slide images. Building upon the original dual-curriculum paradigm, the present framework incorporates refinements in multi-magnification instance representation learning and a normal-control-aware triple-tier contrastive Cox strategy. By progressively learning prognosis-related representations and adaptively aggregating informative instances, DCMIL enables survival prediction from gigapixel WSIs without requiring dense annotations. Extensive experiments across twelve cancer types demonstrate competitive prognostic performance, while additional ablation and visualization analyses support the contributions of the learning components and provide further insights into prognosis-salient morphological patterns.

\textbf{Limitations.} Despite these findings, several challenges remain before broader clinical application. {Although the present study extends the preliminary NeurIPS 2022 framework through methodological refinements and broader experimental validation, a direct comparison of prognostic performance between the original and extended versions of DCMIL under identical experimental settings has not been conducted.} The biological significance of the identified salient regions and uncertainty estimates requires further pathological and molecular validation. In addition, variations in patient populations, tissue processing, image acquisition, and model configurations may affect generalizability across clinical environments. Future research should therefore focus on prospective multi-center validation, standardized preprocessing, and pathologist-guided interpretation. Further integration of histopathological representations with genomic, transcriptomic, and clinical information may enhance the characterization of tumor heterogeneity and improve the reliability of prognosis prediction.

\section*{Data and code availability}
All 12 cancer datasets from The Cancer Genome Atlas Program (TCGA), including WSIs and their corresponding clinical data, are publicly available at https://portal.gdc.cancer.gov/. All codes have been made publicly accessible at
https://github.com/tuuuc/DCMIL.

\section*{Authors' contributions}
Z.N. and Y.Z. conceived the work’s conception and design. Z.N. and C.T. designed the model and trained in its use. C.T. provided the coding, implementation, and technical support. C.T. acquired the data. Z.N. and C.T. analyzed and interpreted the data. Y.Z., Z.N., and C.T. drafted the paper and the figures. Z.W. and F.L. contributed to supplementary experiments, data analysis, and manuscript revision. K.H., J.Z., and Q.F. contributed equally as co-advisers. All authors critically reviewed the paper and the results and approved the final version.

\section*{Declaration of interests}
All  authors declare no competing
interests.

\section*{Acknowledgements}
This work was supported in part by the National Natural Science Foundation of China under Grant U22A20350 and Grant 61971213, in part by the Basic and Applied Basic Research Foundation of Guangdong Province under Grant 2024A1515012004, in part by the Guangdong Provincial Key Laboratory of Medical Image Processing under Grant No.2020B1212060039, and in part by the Special Funds for the Cultivation of Guangdong College Students' Scientific and Technological Innovation under Grant No.pdjh2024a087 and Grant No.pdjh2023a0099.

\bibliographystyle{elsarticle-num} 
\bibliography{reference}

\newpage
\appendix
\section{Main notations used in the work}
In this part, we list the main notations in Table \ref{table_notation} for clear reference

\section{Pseudocode of the Proposed Method}
The detailed procedure of the proposed method is summarized in \textbf{Algorithm 1}.

\section{Implementation Details.}
We implemented all competing methods using the Pytorch 2.0 library on Python 3.9. All intensive calculations were offloaded to a 32 GB NVIDIA Tesla V100 GPU.

For Curriculum I, 
{we trained the multi-magnification instance encoder using weak risk-stratification supervision. Specifically, patient-level survival outcomes were first converted into high- and low-risk labels according to a predefined survival threshold, and the obtained patient-level risk labels were assigned to their corresponding tiles for weakly supervised instance learning.}
We adopted Adam optimizer for network training.
The learning rate and batch size were set to $10^{-5}$ and $12$, respectively.
The weight coefficients $\beta_{\Omega}$ and $\beta_{R}$ were set to $10^{-5}$ and $1$, respectively.
To better guide fine-grained representation learning with coarse-grained saliency map, we first pre-trained the \textit{1}-st (i.e., 5$\times$) branch for 200 epochs. 
Subsequently, its parameters were shared with the \textit{2}-nd (i.e., 10$\times$) and \textit{3}-rd (i.e., 20$\times$) branches, and all of them were trained together for 50 epochs with the early-stopping patience of 5 epochs.
We selected a three-year survival time as the threshold for six datasets (BLCA, COAD, HNSC, LIHC, OV, and STAD) and a five-year survival time for the remaining datasets (BRCA, KIRC, LUAD, LUSC, THCA, and UCEC).
The reasons for these choices are twofold: 
1) Three-year and five-year survival times are significant clinical prognosis assessment timepoints. 
2) Selecting three-year or five-year survival time as the threshold ensures a relatively balanced distribution of positive and negative samples, which benefits effective model training.

For Curriculum II, 
{the instance encoder learned from Curriculum I was frozen to provide stable prognosis-aware tile representations, while the soft-bag selection module, constrained self-attention module, and survival prediction head were trained for patient-level prognosis inference under Cox supervision and contrastive regularization, thereby mitigating potential bias from the surrogate risk stratification task.}
In this stage, we adopted SGD optimizer with the momentum of 0.9 for network training.
The learning rate, batch size, and epoch number were set to $10^{-5}$, 64 and 1000, respectively.
The soft-bag size $N_B$ was set to 30 and the weight coefficients $\beta_{tcl}$, $\beta_{adc}$, and $\beta_{s}$ were set to $1$, $0.1$, and $10^{-4}$, respectively.

{
\section{Computational Complexity and Matched Efficiency Analysis}

\begin{table}[ht]
\centering
\caption{\textbf{Matched computational comparison between the current DCMIL and the preliminary NeurIPS 2022 implementation.}
Both methods were profiled under the same BLCA fold and hardware environment. FLOPs correspond to the profiled workload under the specified stage.}
\label{tab:computational_efficiency}
\resizebox{\textwidth}{!}{
\begin{tabular}{llcc}
\toprule
\textbf{Stage} & \textbf{Metric} & \textbf{NeurIPS 2022} & \textbf{Current DCMIL} \\
\midrule
Curriculum I
& Parameters (M) & 9.660 & 222.557 \\
& FLOPs (GFLOPs) & 752.056 & 11584.468 \\
& Training time / batch (s) & 0.417 & 0.361 \\
& Total training time (s) & 2670.79 & 11860.15 \\
& Tumor-WSI inference time (s/WSI) & 0.589 & 9.760 \\
& Peak allocated GPU memory (GB) & 12.940 & 13.078 \\
\midrule
Curriculum II
& Trainable parameters (M) & 0.184 & 0.161 \\
& Total training time (s) & 59.20 & 68.11 \\
\midrule
Full pipeline
& Total C-I+C-II training time (h) & 0.76 & 3.31 \\
\bottomrule
\end{tabular}}
\end{table}

To provide a quantitative assessment of computational complexity, we further performed a matched profiling analysis between the current DCMIL framework and our preliminary NeurIPS 2022 implementation. Both methods were evaluated using the same BLCA fold, hardware environment, and profiling protocol. We measured model parameters, FLOPs, total training time, WSI-level inference time, throughput, and peak GPU memory for the two curricula separately.

As summarized in Table~\ref{tab:computational_efficiency}, the expanded Curriculum-I encoder in the current DCMIL has a substantially larger modeling capacity, with 222.557M parameters compared with 9.660M in the preliminary version. This is accompanied by an increase in the profiled computational workload from 752.056 to 11,584.468 GFLOPs. The total Curriculum-I training time increased from 2,670.79 s (0.74 h) to 11,860.15 s (3.29 h), while the average tumor-WSI inference time increased from 0.59 to 9.76 s. In contrast, peak allocated GPU memory during Curriculum-I training remained comparable (12.94 vs. 13.08 GB).

Curriculum II remains lightweight relative to the instance-encoding stage. The current implementation contains 0.161M trainable parameters, slightly fewer than the 0.184M parameters in the preliminary version, while complete Curriculum-II training required 68.11 s compared with 59.20 s. Overall, the complete C-I+C-II training pipeline required approximately 3.31 h for the current DCMIL and 0.76 h for the preliminary version.

Overall, the matched profiling results show that the expanded DCMIL framework increases computational requirements primarily in Curriculum I, whereas Curriculum II remains comparatively lightweight. These findings characterize the performance--complexity trade-off introduced by the expanded multi-magnification representation-learning framework.}

\section{Estimating the Mutual Information with the Triple Loss}
As shown in Section 2.4, the optimal value for $\mathbb{L}(\mathbf{B}_n, \widetilde{\mathbf{B}}_n,\widehat{\mathbf{B}}_n, \overline{\mathbf{B}}_n)$ is given by
{ \small
\begin{equation}
    \mathbb{L}(\mathbf{B}_n,\widetilde{\mathbf{B}}_n,\widehat{\mathbf{B}}_n,\overline{\mathbf{B}}_n)=\frac{{p(\mathbf{B}_n|\widetilde{\mathbf{B}}_n)}{p(\mathbf{B}_n|\widehat{\mathbf{B}}_n)}{p(\mathbf{B}_n|\overline{\mathbf{B}}_n)}}{p(\mathbf{B}_n)p(\mathbf{B}_n)p(\mathbf{B}_n)}.
\end{equation}}
Inserting this back into $\mathcal{L}_{tcl}$ and splitting $\mathcal{B}$ into the positive bag and negative bags results in:
{ \footnotesize
\begin{align}
	\mathcal{L}_{tcl}&=-\underset{\mathcal{B}}{\boldsymbol{E}}\left[\log\frac{\mathbb{L}(\mathbf{B}_n,\widetilde{\mathbf{B}}_n,\widehat{\mathbf{B}}_n,{\overline{\mathbf{B}}_n})}{{{\mathbb{L}(\mathbf{B}_n,\widetilde{\mathbf{B}}_n,\widehat{\mathbf{B}}_n,\overline{\mathbf{B}}_n)}+\sum_{i\in{[N]},i\neq n}\mathbb{L}(\mathbf{B}_i,\underline{\mathbf{B}}, \widehat{\mathbf{B}}_n,\overline{\mathbf{B}}_n)}}\right]\\
	&=\underset{\mathcal{B}}{\boldsymbol{E}}\log\left[1 + \frac{1}{\mathbb{L}(\mathbf{B}_n,\widetilde{\mathbf{B}}_n,\widehat{\mathbf{B}}_n,{\overline{\mathbf{B}}_n})}{\sum_{i\in{[N]},i\neq n}\mathbb{L}(\mathbf{B}_i,\underline{\mathbf{B}}, \widehat{\mathbf{B}}_n,\overline{\mathbf{B}}_n)}\right]\\
	&\approx \underset{\mathcal{B}}{\boldsymbol{E}}\log\left[1 + \frac{1}{\mathbb{L}(\mathbf{B}_n,\widetilde{\mathbf{B}}_n,\widehat{\mathbf{B}}_n,{\overline{\mathbf{B}}_n})}{(N-1)\boldsymbol{E}_{i\in{[N]},i\neq n}\mathbb{L}(\mathbf{B}_i,\underline{\mathbf{B}}, \widehat{\mathbf{B}}_n,\overline{\mathbf{B}}_n)}\right] \label{a4} \\
	&=\underset{\mathcal{B}}{\boldsymbol{E}}\log\left[1 + \frac{(N-1)p(\mathbf{B}_n)^3}{{p(\mathbf{B}_n| \widetilde{\mathbf{B}}_n)}{p(\mathbf{B}_n| \widehat{\mathbf{B}}_n)}{p(\mathbf{B}_n|\overline{\mathbf{B}}_n)}}{\boldsymbol{E}_{i\in{[N]},i\neq n}\frac{{p(\mathbf{B}_i| \underline{\mathbf{B}})}{p(\mathbf{B}_i| \widehat{\mathbf{B}}_n)}{p(\mathbf{B}_i|\overline{\mathbf{B}}_n)}}{p(\mathbf{B}_i)^3}}\right]\\
	&=\underset{\mathcal{B}}{\boldsymbol{E}}\log\left[1 + \frac{(N-1)p(\mathbf{B}_n)^3}{{p(\mathbf{B}_n| \widetilde{\mathbf{B}}_n)}{p(\mathbf{B}_n| \widehat{\mathbf{B}}_n)}{p(\mathbf{B}_n|\overline{\mathbf{B}}_n)}}\right]\\
	&\geq \underset{\mathcal{B}}{\boldsymbol{E}}\log\left[N \frac{p(\mathbf{B}_n)^3}{{p(\mathbf{B}_n| \widetilde{\mathbf{B}}_n)}{p(\mathbf{B}_n| \widehat{\mathbf{B}}_n)}{p(\mathbf{B}_n|\overline{\mathbf{B}}_n)}}\right]\\
	&=-\underset{\mathcal{B}}{\boldsymbol{E}}\log\frac{p(\mathbf{B}_n| \widetilde{\mathbf{B}}_n)}{p(\mathbf{B}_n)}- \underset{\mathcal{B}}{\boldsymbol{E}}\log\frac{p(\mathbf{B}_n| \overline{\mathbf{B}}_n)}{p(\mathbf{B}_n)}- \underset{\mathcal{B}}{\boldsymbol{E}}\log\frac{p(\mathbf{B}_n| \widehat{\mathbf{B}}_n)}{p(\mathbf{B}_n)}+\log N\\
	&=-\mathcal{I}(\mathbf{B}_n,{\widetilde{\mathbf{B}}_n})-\mathcal{I}(\mathbf{B}_n,{\overline{\mathbf{B}}_n})-\mathcal{I}(\mathbf{B}_n,{\widehat{\mathbf{B}}_n})+\log N
\end{align}}
Therefore, $\mathcal{I}(\mathbf{B}_n,{\widetilde{\mathbf{B}}_n})+\mathcal{I}(\mathbf{B}_n,{\overline{\mathbf{B}}_n})+\mathcal{I}(\mathbf{B}_n,{\widehat{\mathbf{B}}_n})\geq- \mathcal{L}_{tcl}+\log N$. This relationship also holds for other functions $f$ that result in a higher loss value $\mathcal{L}_{tcl}$. Equation~\ref{a4} becomes more accurate as $N$ increases. Meanwhile, the value of $log(N)-\mathcal{L}_{tcl}$ also increases with larger $N$. Therefore, it is beneficial to use large values of N. 

{\section{External Validation on CPTAC-HNSCC Dataset}

To further evaluate the generalization ability of the proposed method, we conducted an external validation on the independent CPTAC-HNSCC dataset. This dataset was not used during model training or hyperparameter selection, and thus provides an unbiased assessment of model robustness under cross-cohort distribution shift.

We directly applied the model trained on the TCGA-HNSC cohort to the CPTAC-HNSCC dataset without any fine-tuning. The performance was compared against the two best-performing baseline models on the HNSC dataset in the main experiments, namely AMISL and GigaPath.

As shown in Table~\ref{tab_external_hnscc}, the proposed method achieved numerically higher C-index values than both competing methods in both internal HNSC and external CPTAC-HNSCC evaluation.

\begin{table}[h]
\centering
\caption{\textbf{External validation results on CPTAC-HNSCC dataset and comparison on HNSC.}}
\label{tab_external_hnscc}
\begin{tabular}{l c c}
\toprule
Method & HNSC (TCGA) C-index & CPTAC-HNSCC C-index \\
\midrule
AMISL & 0.624$\pm$0.053 & 0.648$\pm$0.070 \\
GigaPath & 0.631$\pm$0.003 & 0.610$\pm$0.025 \\
\textbf{DCMIL (Ours)} & \textbf{0.667$\pm$0.016} & \textbf{0.671$\pm$0.015} \\
\bottomrule
\end{tabular}
\end{table}

\section{Controlled Study of Triple-Tier Contrastive Learning}

To further investigate the contribution of the triple-tier contrastive learning
(TCL) objective, we performed a controlled ablation study under a unified
training and evaluation protocol. The proposed contrastive formulation
regularizes WSI representations through three complementary relationships:
(1) tumor--normal contrast (TN), which encourages discrimination between
tumor and adjacent-normal tissues; (2) risk-level contrast (RL), which
promotes separation between subjects with different survival-derived risk
status; and (3) intra-bag saliency contrast (IS), which models the relationship
between selected high-salient and discarded low-salient regions within
the same WSI bag.

To disentangle the contribution of these relationships, we evaluated eight
configurations:

\begin{itemize}
    \item \textbf{Baseline (w/o CL):} survival prediction without contrastive
    regularization;

    \item \textbf{Tumor--Normal Contrast (TN):} contrastive learning between
    tumor and adjacent-normal tissue representations;

    \item \textbf{Risk-Level Contrast (RL):} contrastive learning between
    subjects with different risk-stratification status;

    \item \textbf{Intra-Bag Saliency Contrast (IS):} contrastive learning
    between selected high-salient and discarded low-salient regions
    within the same WSI bag;

    \item \textbf{TN + RL:} joint tumor--normal and risk-level contrastive
    learning;

    \item \textbf{TN + IS:} joint tumor--normal and intra-bag saliency
    contrastive learning;

    \item \textbf{RL + IS:} joint risk-level and intra-bag saliency
    contrastive learning;

    \item \textbf{Full TCL (TN + RL + IS):} the complete triple-tier
    contrastive formulation incorporating all three relationships.
\end{itemize}

\begin{table}[ht]
\centering
\caption{\textbf{Ablation study of triple-tier contrastive learning across 12 datasets (C-index).}}
\label{tab:tcl_ablation}
\resizebox{\textwidth}{!}{
\begin{tabular}{lcccccccccccc}
\toprule
Method & KIRC & LUAD & LUSC & STAD & THCA & HNSC & LIHC & BRCA & BLCA & COAD & OV & UCEC \\
\midrule

Baseline (w/o CL)
& 0.639 & 0.623 & 0.614 & 0.610 & 0.781 & 0.605 & 0.620 & 0.634 & 0.583 & 0.665 & 0.594 & 0.651 \\

Tumor--Normal Contrast (TN)
& 0.653 & 0.586 & 0.611 & 0.623 & 0.790 & 0.632 & 0.662 & 0.676 & 0.593 & 0.673 & 0.607 & 0.662 \\

Risk-Level Contrast (RL)
& 0.653 & 0.625 & 0.624 & 0.640 & 0.835 & 0.633 & 0.663 & 0.704 & 0.621 & 0.687 & 0.606 & 0.669 \\

Intra-Bag Saliency Contrast (IS)
& 0.652 & 0.613 & 0.635 & 0.646 & 0.835 & 0.633 & 0.663 & 0.704 & 0.590 & 0.667 & 0.613 & 0.663 \\

TN + RL
& 0.671 & 0.648 & 0.616 & 0.651 & 0.847 & 0.640 & 0.676 & 0.710 & 0.612 & 0.700 & 0.595 & 0.688 \\

TN + IS
& 0.662 & 0.619 & 0.662 & 0.655 & 0.846 & 0.642 & 0.683 & 0.712 & 0.632 & 0.712 & 0.626 & 0.650 \\

RL + IS
& 0.688 & 0.636 & 0.664 & 0.659 & 0.845 & 0.638 & 0.692 & 0.706 & 0.611 & 0.692 & 0.616 & 0.656 \\

Full TCL (TN + RL + IS)
& \textbf{0.697} & \textbf{0.655} & \textbf{0.671} & \textbf{0.683} & \textbf{0.848} & \textbf{0.647} & \textbf{0.695} & \textbf{0.712} & \textbf{0.653} & \textbf{0.720} & \textbf{0.628} & \textbf{0.713} \\

\bottomrule
\end{tabular}
}
\end{table}

The results show that the three contrastive relationships provide complementary constraints on representation learning. TN introduces normal tissue as an additional reference distribution and encourages discrimination between tumor and non-tumor morphological representations. RL aligns representation learning with survival-related risk differences across subjects, whereas IS preserves the relationship between selected and discarded regions within the same WSI bag. Overall, combining multiple relationships generally provides more consistent performance than using individual contrastive objectives alone, and the full TN+RL+IS formulation achieves the strongest overall performance across the evaluated cohorts.

{To further assess the contribution of tumor--normal contrast, we compared the RL+IS and TN+RL+IS configurations under the same experimental protocol. Across the 12 cancer cohorts, the RL+IS configuration achieved a mean C-index of 0.675, whereas the complete triple-tier formulation achieved 0.694, corresponding to a mean absolute difference of approximately 0.018. The full configuration yielded higher C-index values across all evaluated cohorts, suggesting that tumor--normal contrast provides complementary information beyond the existing risk-level and intra-bag saliency relationships. This controlled comparison quantifies the incremental contribution of tumor--normal contrast to prognostic performance while holding the remaining model components and evaluation protocol fixed.
}
\section{End-to-End Analysis of Dual-Curriculum Design}

The dual-curriculum framework is motivated by an easy-to-hard learning paradigm, where instance-level representations learned in the first curriculum (C-I) are subsequently transferred to the second curriculum (C-II) for survival prediction. A potential concern is that the use of frozen instance representations from C-I may introduce error propagation, since the surrogate risk stratification task in C-I is not directly optimized for survival prediction.

To investigate this issue, we further evaluate an end-to-end variant of the proposed framework. In this setting, the instance encoder in C-I is jointly optimized with the survival objective in C-II, rather than being frozen after C-I. Specifically, gradients derived from the bag-level survival loss are propagated back to C-I, enabling full fine-tuning of the instance encoder under survival supervision.

We compare the original frozen dual-curriculum strategy (DCMIL) with the end-to-end variant across 12 datasets. 
As shown in Table~\ref{tab:e2e}, the frozen two-stage strategy achieved numerically higher C-index values than the end-to-end variant across all 12 datasets. These results suggest that freezing the Curriculum-I representations does not necessarily compromise downstream survival prediction performance and may provide a stable optimization strategy.

\begin{table}[h]

\centering

\caption{\textbf{Comparison between frozen dual-curriculum (DCMIL) and end-to-end fine-tuning.} End-to-end results are reported only on datasets where full training convergence is achieved.}

\label{tab:e2e}

\setlength{\tabcolsep}{5pt}

\resizebox{\textwidth}{!}{

\begin{tabular}{lcccccccccccc}
\toprule
Method & KIRC & LUAD & LUSC & STAD & THCA & HNSC & LIHC & BRCA & BLCA & COAD & OV & UCEC \\
\midrule
End-to-End
& 0.667$_{\scriptsize \pm .082}$
& 0.642$_{\scriptsize \pm .028}$
& 0.643$_{\scriptsize \pm .020}$
& 0.631$_{\scriptsize \pm .020}$
& 0.860$_{\scriptsize \pm .045}$
& 0.621$_{\scriptsize \pm .019}$
& 0.678$_{\scriptsize \pm .051}$
& 0.660$_{\scriptsize \pm .008}$
& 0.635$_{\scriptsize \pm .027}$
& 0.709$_{\scriptsize \pm .052}$
& 0.566$_{\scriptsize \pm .011}$
& 0.647$_{\scriptsize \pm .039}$ \\
\midrule
DCMIL
& 0.676$_{\scriptsize \pm .018}$
& 0.646$_{\scriptsize \pm .020}$
& 0.659$_{\scriptsize \pm .024}$
& 0.672$_{\scriptsize \pm .017}$
& 0.871$_{\scriptsize \pm .052}$
& 0.667$_{\scriptsize \pm .016}$
& 0.681$_{\scriptsize \pm .022}$
& 0.715$_{\scriptsize \pm .017}$
& 0.669$_{\scriptsize \pm .047}$
& 0.718$_{\scriptsize \pm .030}$
& 0.616$_{\scriptsize \pm .022}$
& 0.726$_{\scriptsize \pm .035}$ \\
\bottomrule
\end{tabular}
}
\end{table}

\section{Additional Analysis on Risk Stratification Threshold Selection}
\label{sec:threshold_appendix}

To address the concern regarding potential bias introduced by dataset-specific survival threshold selection, we conducted a comprehensive sensitivity analysis across multiple clinically meaningful time horizons, including 1-year, 3-year, 5-year, and 10-year thresholds. For each dataset and threshold setting, patients were dichotomized into high-risk and low-risk groups according to survival status relative to the selected cutoff. To ensure statistical reliability, we excluded cases where either risk group contained fewer than five samples. Table~\ref{tab:threshold_sensitivity} reports the corresponding C-index and the proportion of high-risk cases under all evaluated settings.

The results demonstrate that model performance remains relatively stable across different threshold choices, indicating that the proposed method is not overly sensitive to a specific survival cutoff. Among all settings, the 3-year and 5-year thresholds consistently achieve the best or near-best C-index across most datasets, while also maintaining more balanced risk group distributions. In contrast, the 1-year threshold is often affected by severe class imbalance due to insufficient event accumulation, whereas the 10-year threshold suffers from increased censoring effects in several cancer types, both of which may weaken the reliability of risk stratification.

We further observe that the optimal threshold exhibits dataset-dependent heterogeneity. For rapidly progressing tumors such as LIHC and STAD, shorter horizons tend to provide more discriminative stratification, whereas for relatively indolent cancers such as BRCA and THCA, longer horizons may better reflect long-term survival differences. Despite these variations, the 3-year and 5-year thresholds provide a robust compromise across heterogeneous datasets in terms of both predictive performance and risk group balance.

Overall, these findings indicate that while survival threshold selection is inherently dataset-dependent, the 3-year and 5-year settings offer a stable and well-balanced choice across diverse cancer types. Importantly, the observed performance improvements of our method are consistent across all threshold settings, suggesting that the reported results are not an artifact of a specific cutoff design.

\begin{table}[t]
\centering
\caption{\textbf{Sensitivity analysis of survival time threshold selection.}
Each threshold reports C-index and high-risk ratio (\%). Underlined values indicate the best C-index across thresholds for each dataset.}
\label{tab:threshold_sensitivity}
\setlength{\tabcolsep}{4pt}
\resizebox{\textwidth}{!}{
\begin{tabular}{lcccccc|ccccccc}
\toprule
& \multicolumn{6}{c|}{\textbf{3-year threshold group}} 
& \multicolumn{6}{c}{\textbf{5-year threshold group}} \\
\cmidrule(lr){2-7} \cmidrule(lr){8-13}

\textbf{Threshold}
& BLCA & COAD & HNSC & LIHC & OV & STAD
& BRCA & KIRC & LUAD & LUSC & THCA & UCEC & \textbf{Avg.} \\
\midrule

\multirow{1}{*}{\textbf{1-year C-index}}
& 0.625 & 0.676 & 0.638 & 0.655 & 0.559 & 0.618
& 0.608 & 0.681 & 0.636 & 0.601 & - & 0.692 & \textbf{0.635} \\
\multirow{1}{*}{High-risk Ratio(\%)}
& 20.3 & 10.4 & 17.7 & 14.5 & 9.00 & 23.0
& 1.80 & 10.4 & 11.8 & 13.7 & 0.670 & 4.53 & \textbf{11.4} \\
\midrule

\multirow{1}{*}{\textbf{3-year C-index}}
& \underline{0.630} & \underline{0.701} & \underline{0.647} & \underline{0.659} & \underline{0.581} & \underline{0.640}
& 0.655 & 0.683 & 0.637 & 0.611 & 0.848 & 0.704 & \textbf{0.665} \\
\multirow{1}{*}{High-risk Ratio(\%)}
& 64.1 & 40.5 & 57.1 & 50.8 & 39.4 & 74.6
& 13.7 & 27.9 & 48.7 & 49.4 & 5.19 & 24.4 & \textbf{41.3} \\
\midrule

\multirow{1}{*}{\textbf{5-year C-index}}
& 0.627 & 0.698 & 0.646 & 0.656 & 0.574 & 0.627
& \underline{0.687} & \underline{0.689} & \underline{0.637} & \underline{0.633} & \underline{0.912} & \underline{0.713} & \textbf{0.675} \\
\multirow{1}{*}{High-risk Ratio(\%)}
& 78.0 & 66.1 & 78.7 & 72.8 & 70.2 & 90.1
& 28.3 & 52.6 & 75.9 & 67.5 & 13.6 & 41.9 & \textbf{61.3} \\
\midrule

\multirow{1}{*}{\textbf{10-year C-index}}
& 0.608 & 0.671 & 0.607 & - & 0.545 & -
& 0.641 & 0.658 & 0.615 & 0.607 & 0.880 & 0.688 & \textbf{0.652} \\
\multirow{1}{*}{High-risk Ratio(\%)}
& 96.7 & 93.0 & 96.3 & 99.2 & 96.6 & 99.3
& 76.0 & 92.6 & 94.5 & 92.6 & 47.1 & 92.5 & \textbf{89.7} \\

\bottomrule
\end{tabular}
}
\end{table}

\section{Hyperparameter Sensitivity Analysis and Limitations}
\label{sec:sensitivity_limitation}

We systematically investigated the sensitivity of the proposed model to seven key hyperparameters, including the number of selected regions $N_B$, the loss weights $\beta_{tcl}$, $\beta_{adc}$, and $\beta_s$, the activation threshold $\iota$, the Gumbel--Softmax temperature $\tau$, and the coefficient $\eta$ in Eq.~(13). All hyperparameters were determined according to validation-set performance and were fixed before evaluation on the test set. The corresponding heatmaps and sensitivity curves are presented in Fig.~\ref{fig:parameter_analysis}.

The heatmaps illustrate the model performance under different settings of $N_B$, $\beta_{tcl}$, $\beta_{adc}$, and $\beta_s$. Stable high-performance regions can be observed around the selected configuration, indicating that the performance is not dependent on a single isolated parameter setting. A small $N_B$ may discard diagnostically relevant regions and provide an insufficient representation of tissue heterogeneity, whereas an excessively large $N_B$ may introduce redundant or weakly relevant regions and increase the computational cost. Similarly, overly small loss weights weaken the effects of their corresponding learning objectives, while overly large weights may cause individual objectives to dominate the joint optimization. Based on the validation results, we selected
\[
N_B=30,\qquad
\beta_{tcl}=1,\qquad
\beta_{adc}=0.1,\qquad
\beta_s=10^{-4}.
\]

The sensitivity curves further demonstrate the effects of $\iota$, $\tau$, and $\eta$. For the activation threshold $\iota$, a small value retains more weak or noisy GradCAM responses, whereas an excessively large value may remove informative regions. The model achieves the best balance between suppressing noisy activations and preserving relevant information at $\iota=0.4$. For the Gumbel--Softmax approximation, several candidate temperature values were evaluated:
\[
\tau\in\{0.1, 0.5, 1, 2, 5\}.
\]
A relatively large temperature produces a smoother and more diffuse probability distribution, facilitating gradient-based optimization but providing a less distinct approximation to discrete selection. In contrast, a very small temperature generates sharper and sparser selection probabilities but may cause unstable gradients and premature hard selection. It should be noted that $N_B$ determines the final number of retained regions; therefore, $\tau$ affects the sparsity and confidence of the soft selection probabilities during training rather than changing the final top-$N_B$ selection cardinality. The best validation performance is obtained at $\tau=1$, which provides an appropriate balance between optimization stability and selection sharpness.

We also evaluated different values of $\eta$:
\[
\eta\in\{10^{-4}, 5\times10^{-4}, 10^{-3}, 5\times10^{-3}\}.
\]
A very small $\eta$ weakens the contribution of the corresponding constraint in Eq.~(13), whereas an excessively large value overemphasizes this constraint and interferes with the optimization of the primary objective. The model achieves its best validation performance at $\eta=10^{-3}$, providing an empirical justification for the adopted setting. Overall, the heatmaps and sensitivity curves show that the model maintains relatively stable performance within reasonable ranges of the seven hyperparameters. 

Despite the encouraging results, this study has several limitations. 
{First, external validation is currently limited to the CPTAC-HNSCC cohort. Although this experiment provides evidence of cross-cohort robustness for HNSC, independent external validation remains necessary for the other evaluated cancer types and across broader prospective multi-center settings.}
Differences in scanners, staining protocols, tissue preparation procedures, patient populations, and clinical settings may result in performance variations across institutions. Future studies will therefore evaluate the method using independent and prospective multi-center cohorts. Second, the demographic and clinical composition of TCGA may introduce population biases because certain racial, ethnic, geographic, and clinical subgroups are underrepresented. Additional subgroup analyses and fairness assessments on more diverse populations are necessary to determine whether the model performs consistently across different patient groups.

Although the sensitivity analysis indicates reasonable robustness, the model still depends on several key hyperparameters, which may require recalibration when applied to new cohorts. Moreover, the region-selection procedure relies on GradCAM, introducing additional computational cost because gradients must be calculated for the target feature maps. GradCAM activation maps may also vary across model checkpoints or under small input perturbations, potentially affecting the stability of the selected regions. More efficient and stable region-localization strategies will be investigated in future work. Finally, clinical deployment requires standardized preprocessing and quality-control procedures, external and prospective validation, subgroup fairness evaluation, uncertainty estimation, and integration with existing pathological diagnostic workflows. The proposed model is intended to serve as a decision-support system that highlights potentially informative regions and provides case-level predictions for review by pathologists, rather than replacing clinical specialists.

\begin{figure*}[t!]
    \renewcommand\thefigure{S\arabic{figure}}
    \centering
    \includegraphics[width=0.9\textwidth]{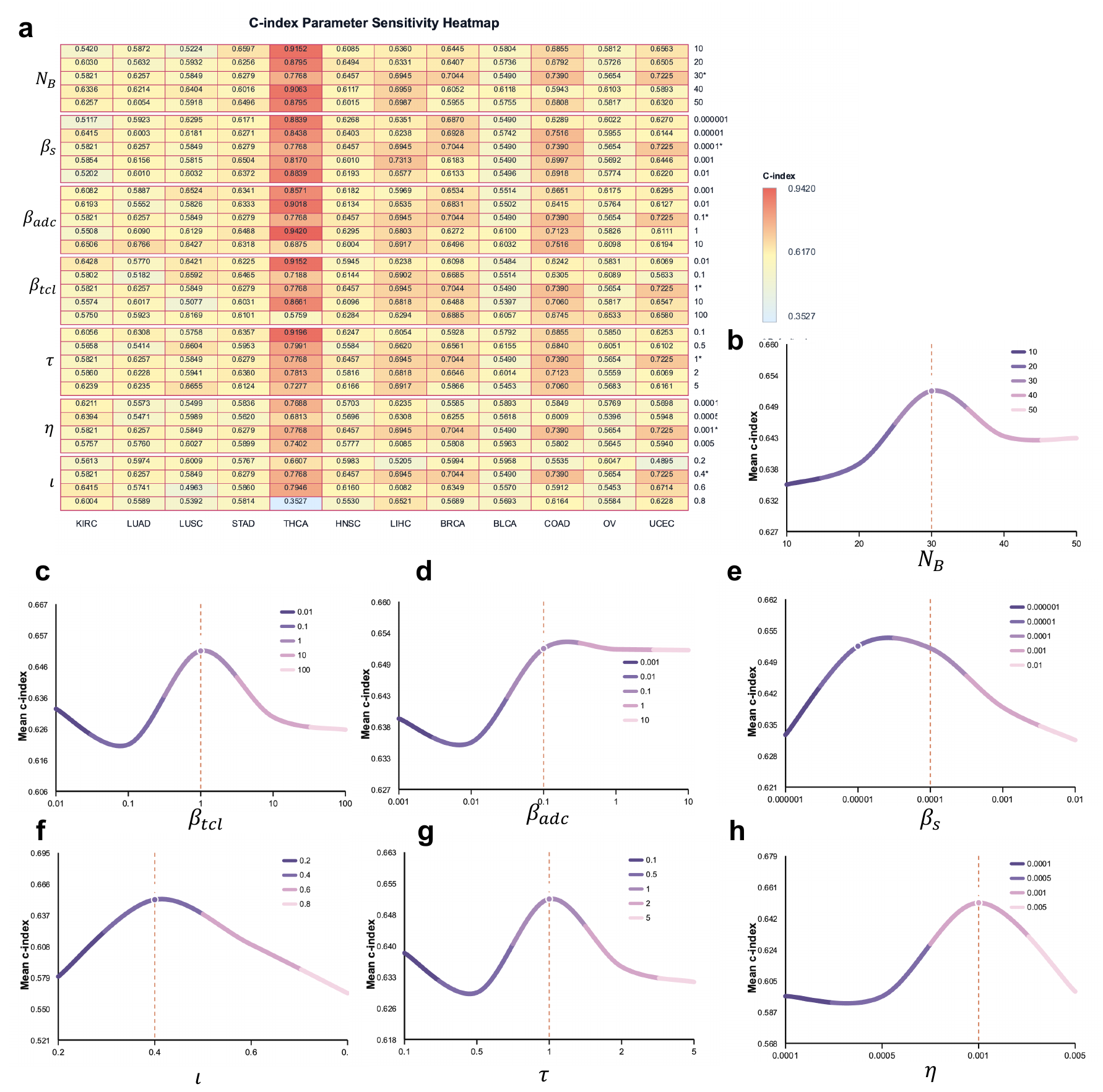}
    \caption{\textbf{Comprehensive hyperparameter sensitivity analysis.} Panels (a) show dataset-wise heatmaps illustrating the variation of C-index under different hyperparameter settings. Panels (b--h) present the corresponding averaged C-index trends across all datasets for each hyperparameter, including $N_B$, $\beta_{tcl}$, $\beta_{adc}$, $\beta_{s}$, threshold $\iota$, Gumbel-Softmax temperature schedule, and regularization coefficient $\eta$. These results demonstrate both dataset-specific sensitivity patterns and overall performance stability, supporting the robustness of the proposed framework and the selected hyperparameter configuration.
    }	
    \label{fig:parameter_analysis}
\end{figure*}

\section{Comparison Between MC Dropout and Seed Ensemble}
To further examine the robustness of instance-level uncertainty estimation, we additionally explored a seed ensemble as an alternative to MC Dropout in Fig.~\ref{fig:seed_ensemble}. Specifically, 30 models were independently trained using different random seeds while maintaining identical datasets, data splits, model architectures, and hyperparameters. During inference, the prediction probabilities produced by these models were averaged to obtain the final prediction, while their variation was used as an intuitive indication of instance-level uncertainty.
Compared with the seed ensemble, MC Dropout produced more concentrated and consistent predictions across repeated inference runs in the representative COAD cases, indicating greater stability in instance-level uncertainty estimation. Importantly, both approaches consistently assigned the representative patients to their corresponding high- and low-risk groups. This comparison therefore provides additional qualitative evidence that the observed uncertainty patterns are not specific to a single stochastic inference strategy and confirms the model’s ability to distinguish between patients with different risk levels.

\begin{figure*}[t!]
    \renewcommand\thefigure{S\arabic{figure}}
    \centering
    \includegraphics[width=0.9\textwidth]{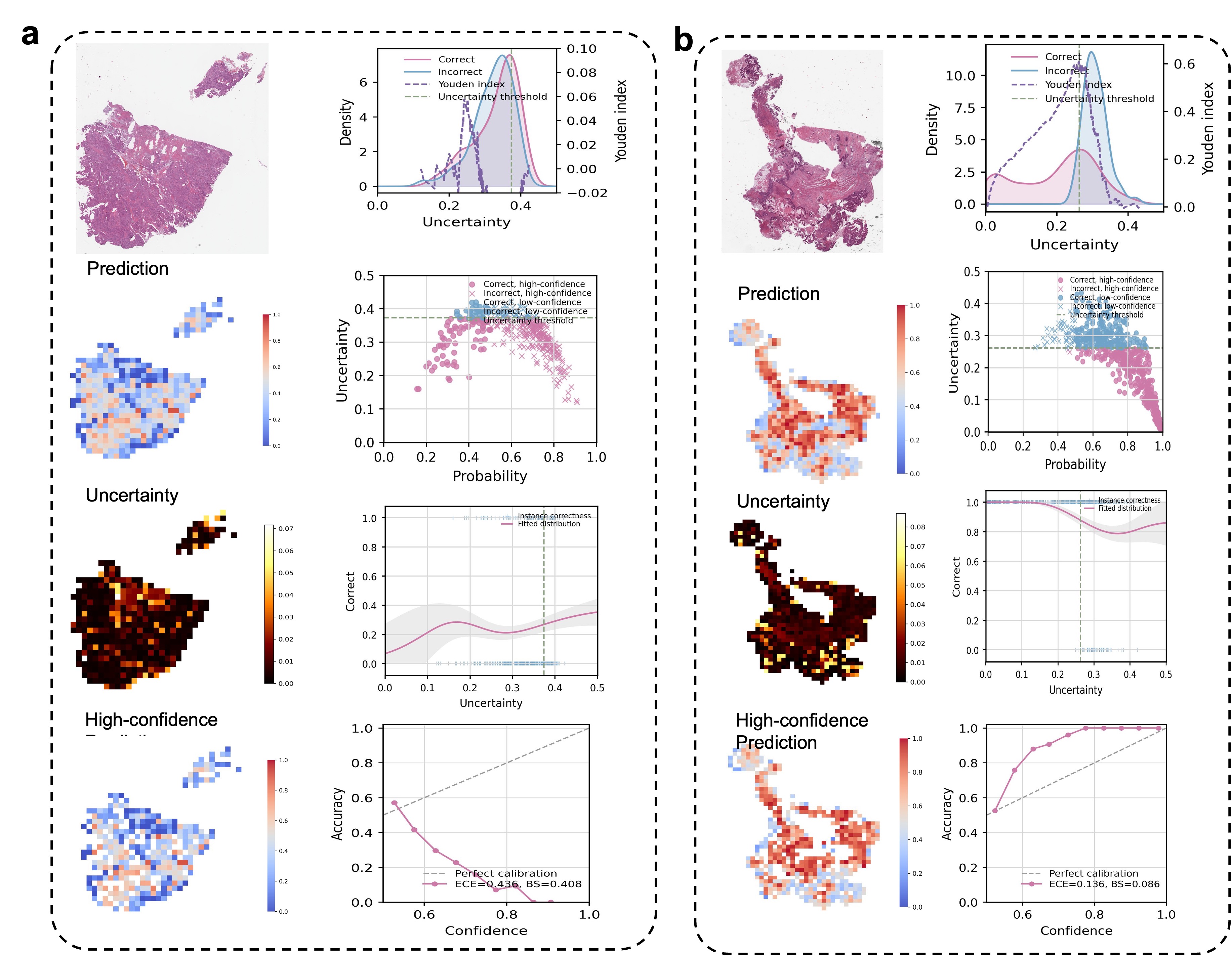}
    \caption{\textbf{Seed ensemble-based uncertainty estimation for two representative cases from the COAD cohort.} (a) A low-risk patient and (b) a high-risk patient. Predictions were obtained from 30 independently trained models with different random seeds; their mean represents the final prediction, while their variability reflects instance-level uncertainty.
    }	
    \label{fig:seed_ensemble}
\end{figure*}
}

\section{Additional Visualization Results}
To further examine the consistency and interpretability of the proposed model across additional cancer types, we present visualization results on the LUAD (Figure~\ref{fig_2_luad}–\ref{fig_6_luad}) and HNSC (Figure~\ref{fig_2_hnsc}–\ref{fig_6_hnsc}) datasets.

\begin{table}[p]
\renewcommand\thetable{S\arabic{table}}
\caption{\textbf{Main notations used in the work.}}
\label{table_notation}
\centering
\renewcommand{\arraystretch}{}
\resizebox{\textwidth}{!}{
	\begin{tabular}{|c|l|l|}
		\hline
		& Symbol & Description \\ 
		\hline\hline
		
		\multirow{5}{*}{\rotatebox{90}{Indices}}
		& $S$ & Number of magnifications (branches) ($s\in\{1,\dots,S\}$) \\ 		
		& $N$ & Number of patients ($n\in\{1,\dots,N\}$) \\ 
		& $N_n$ & Number of instances in the $n$-th bag \\ 		
		& $N_B$ & Number of selected instances in each bag \\ 		
		& {{}$C,D,D_B,\widehat{D}$} & {{}Feature dimension of $\mathbf G_n$, $\mathbf E_n$, $\mathbf B_n$, or $\mathbf{Q}_n/\mathbf{K}_n/\mathbf{V}_n$} \\ 	
		\hline\hline	
		
		\multirow{7}{*}{\rotatebox{90}{Input}} 
		& $\mathbf X_n$ &The $n$-th bag (patient)\\
		& {{}$\mathbf x_{n,i}$} & {{}The $i$-th instance of the $n$-th bag} \\
		& {{}$\mathbf T_n$} & {{}Observed survival time of the $n$-th patient} \\ 		
		& {{}$\mathbf T_r$} & {{}Time threshold} \\
		& {{}$\boldsymbol{\delta}_n$} & {{}Event indicator of the $n$-th patient} \\	
		& $\mathbf Y_n$ & Risk stratification status of the $n$-th patient\\ 
		& {{}$\beta_{\Omega},\beta_{R},\beta_s,\beta_{tcl},\beta_{adc}$} & {{}Weight coefficients in Curriculum I and II} \\
		\hline\hline
		
	\multirow{10}{*}{\rotatebox{90}{Output}} 
		& {{}$\mathbf p_{n,i}^{s}$} & {{}Predicted probability of the $i$-th instance of the $n$-th bag in the $s$-th branch} \\
		& {{}$\widehat{\mathbf T}_n$} & {{}Estimated survival time of the $n$-th patient} \\ 
		& {{}$\mathcal L_{\textrm{I}},\mathcal L_{\textrm{II}}$} & {{}Loss functions for Curriculum I and II} \\		
		& {{}$\mathcal L_{\ell}$} & {{}Empirical loss} \\			
		& {{}$\mathcal L_{\Omega}$} & {{}Structural loss} \\		
		& {{}$\mathcal L_{R}$} & {{}Pairwise ranking loss} \\
		& {{}$\mathcal L_{cox}$} & {{}Cox proportional hazard regression loss} \\
		& {{}$\mathcal L_{tcl}$} & {{}Triple-tier contrastive learning loss} \\		
		& {{}$\mathcal L_{adc}$} & {{}Absolute distance constraint} \\
		& {{}$\mathcal L_{s}$} & {{}Sparseness loss} \\
		\hline\hline
		\multirow{16}{*}{\rotatebox{90}{Feature map}} 
		& {{}$\alpha_{n,i}^{s}$} & {{}Gradient of the score for $\mathbf p_{n,i}^{s}$}\\			
		& {{}$\mathbf m_{n,i}^{s}$} & {{}Salient mask of the $i$-th instance of the $n$-th bag in the $s$-th branch}\\		
		& {{}$\widehat{\mathbf x}^s_{n,i}$} & {{}Highlighted input of the $i$-th instance of the $n$-th bag in the $s$-th branch} \\
		& {{}$\mathbf z_{n,i}$} & {{}Instance token embedding}	\\	
		& {{}$\mathbf g_{n,i}$} & {{}Multi-scale instance representation of the $i$-th instance of the $n$-th bag}	\\	
  		& $\widehat{\mathbf h}_n$ & Indicator that adaptively selects representative instances of the $n$-th bag\\
  		& $\mathbf G_n$ & A set of instance representations of the $n$-th bag\\ 
		& {{}$\mathbf{B}_n$} & {{}Sparse soft-bag representation of the $n$-th bag}\\	
		& {{}$\mathcal B_n^{\mathcal O^+}/\mathcal B_n^{\mathcal O^-}$} & {{}Collection of bags from ${\mathcal{O^+}}$ (or ${\mathcal{O^-}}$) with the same source as $\mathbf{B}_n$}\\
  		& {{}$\mathcal B$} & {{}Collection of all bags}\\
		& {{}$\mathcal B^{\underline{\mathcal{O}}}, \mathcal B_n^{\mathcal O}$} & {{}Collection of bags (expect $\mathbf{B}_n$) from the source ${\underline{\mathcal{O}}}$ or ${\mathcal{O}}$}\\	
		& {{}$\overline{\mathbf{B}}_n$} & {{}Integrated representation of the discarded instances}\\			
		& {{}$\widehat{\mathbf{B}}_n$} & {{}Integrated representation of $B_n^{\mathcal O^+}$ (or $B_n^{\mathcal O^-}$)}\\
		& {{}$\underline{\mathbf{B}}, \widetilde{\mathbf{B}}_n$} & {{}Integrated representations of $\mathcal B^{\underline{\mathcal{O}}}$ or $\mathcal B_n^{\mathcal O}$}\\	
		& {{}$\mathbf{E}_n,\widehat{\mathbf{E}}_n, \overline{\mathbf{E}}_n$} & {{}Representation of the (all, selected, or discarded) instances of the $n$-th bag}\\				
		& $\mathbf{Q}_n,\mathbf{K}_n,\mathbf{V}_n$ & Feature spaces in $\mathbb{D}$\\				
		\hline\hline					
		
		\multirow{13}{*}{\rotatebox{90}{Network components}}
		& $\mathbb{A}$ & {{}Instance aggregation function} \\  
		& $\mathbb{B}$ & Soft-bag learning module \\ 
		& $\mathbb{C}$ & Risk stratification function (i.e, multilayer perceptron (MLP) heads) \\ 	
		&$\mathbb{D}$ & Constrained self-attention module \\									
		& $\mathbb{E}$ & Instance encoding function \\ 
		&$\mathbb{F}$ & Transformer extractor  \\ 
		&$\mathbb{G}$ &  Attention aggregator\\ 				
		&$\mathbb{H}$ & {{}Indicator function} \\ 
		& $\mathbb{K}_1$, $\mathbb{K}_2$ & Bag or instance aggregation networks\\	
		& $\mathbb{L}$ & Log-bilinear function\\			
            & $\mathbb{P}$ & Linear layer\\	
		& $\mathbb{R}$ & "ReLU" activation function \\ 
		& $\mathbb{S}$ & Prognosis inference function \\ 
		\hline\hline
		
		\multirow{6}{*}{\rotatebox{90}{Others}} 	
		& $\phi$ & "Softmax" activation function \\ 	
		& {{}$\boldsymbol \theta^i_{\mathbb{F}^s}$} & {{}Parameter of the $i$-th module in $\mathbb{F}^s$} \\ 		
		&{{} $\mathbf{W}_n,\mathbf{W}_Q/\mathbf{W}_K/\mathbf{W}_V,\mathbf{W}_\mathbb{L}/\mathbf{V}_{\mathbb{L}},\mathbf{W}_{\mathbb{S}}$} & {{}Weight matrices in $\mathbb{H},\mathbb{D},\mathbb{L},\mathbb{S}$}\\	
		& {{}$\mathcal O=\{\mathcal O^+,\mathcal O^-\}$} & {{}Distribution of (high-risk or low-risk) tumor tissue slides} \\
		& {{}$\underline{\mathcal{O}}$}
		& {{}Distribution of normal tissue slides} \\
		& {{}$\iota$,$\eta$,$\kappa$}
		& {{}Hyperparameters in Curriculum I and II} \\
		\hline	
	\end{tabular}
 }
\end{table}	
\newpage

\begin{algorithm}[p]
\scriptsize  

	\caption{{\footnotesize \textbf{Pseudocode of the Proposed Method.}}} 
	\label{algorithm}
 
	\textbf{Input:} Dataset {$\{\mathbf X_n, \mathbf T_n,\boldsymbol{\delta}_n\}^N_{n=1}$}, risk stratification status $\mathbf Y_n=\{\mathbf y_{n,i}^{s}\}_{i=1}^{N_n}$, and the weight coefficients $\beta_{\Omega},\beta_{R},\beta_s,\beta_{tcl},\beta_{adc}$. \\ 
	\textbf{Output:} Prognosis inference $\widehat{\mathbf T}_n\gets\mathbb{S}(\mathbb{A} \{\mathbb{E}(\mathbf x_{n,i}):\mathbf x_{n,i}\in{\mathbf X_n}\})$, where $\{\mathbb{P}^s\}_{s=1}^S$, $\{\mathbb{F}^s\}_{s=1}^S$, and $\{\mathbb{G}^s\}_{s=1}^S$ form $\mathbb{E}$ while $\mathbb{B}$ and $\mathbb{D}$ constitute $\mathbb{A}$.
	
	\textbf{Curriculum I (C-I):} Saliency-guided weakly-supervised instance encoding with cross-scale tiles.
	\begin{algorithmic}
		\State $s\gets 1$
		\While{$s\leq S$}
		\For{$[n,i]=[1,1]\to [N,N_n]$}
		\If{$s==1$}
		\State $\widehat{\mathbf x}_{n,i}^s\gets \mathbf x_{n,i}^s$
		\Else
		\State $\alpha^{s-1}_{n,i}\gets\frac{\partial {\mathbf p_{n,i}^{s-1}}}{\partial \mathbf z_{n,i}^{s-1}}$ \Comment{Obtain token gradient through backward propagation}
		\State${\mathbf m_{n,i}^{s-1}}\gets \mathbb{R}(\alpha^s_{n,i} \mathbf z_{n,i}^{s})\geq \iota$ \Comment{Generate salient mask}
		\State $\widehat{\mathbf x}_{n,i}^s\gets \mathbf m_{n,i}^{s-1}\otimes\mathbf x_{n,i}^s$ \Comment{Utilize salient regions to highlight the input}
		\EndIf
		\State $\mathbf z_{n,i}^{s}\gets\mathbb{F}^{s}(\mathbb{P}^{s}(\widehat{\mathbf x}_{n,i}^{s}))$	\Comment{Generate instance token embedding}
		\State $\mathbf g_{n,i}^{s}\gets \mathbb{G}^{s}(\mathbf z_{n,i}^s,\mathbf g_{n,i}^{s-1})$ \Comment{Obtain instance representation}
		\State $\mathbf p_{n,i}^{s}\gets\mathbb{C}^{s}(\mathbf g_{n,i}^s)$ \Comment{Predict risk stratification status}
        \State $\mathcal{L}_{\ell}\gets$ the empirical loss calculated with $\{\mathbf p_{n,i}^{s},\mathbf y_{n,i}^{s}\}$	
		\State $\mathcal{L}_{R}\gets$ the pairwise ranking loss calculated with $\{\mathbf p_{n,i}^{s-1},\mathbf p_{n,i}^{s},\mathbf y_{n,i}^{s}\}$
		\EndFor			
		\State $\mathcal L_{\Omega}\gets$ the structural loss calculated with $\{\boldsymbol\theta^{i}_{\mathbb{F}^s},\boldsymbol \theta^{i}_{\mathbb{F}^{s-1}}\}_{i=1}^{s-1}$
		\State $\mathcal L_{\textrm{I}}=\mathcal L_{\ell}+\beta_{\Omega} \mathcal L_{\Omega} {+\beta_{R} \mathcal L_{R}}$ \Comment{Hybrid loss of Curriculum I}
		\State Update $\{\mathbb{P}^{s},\mathbb{F}^{s},\mathbb{G}^{s},\mathbb{C}^{s}\}$ by gradient descent	
		\State $s\gets s+1$
		\EndWhile
		\State ${\mathbf g_{n,i}}\gets \mathbf g_{n,i}^S$. \Comment{Obtain multi-scale instance representation} 
	\end{algorithmic}
	\textbf{Curriculum II (C-II):} Contrastive-enhanced Soft-bag Prognosis Inference.
	\begin{algorithmic}
		\For{$n=1\to N$}
		\State $\mathbf G_n\gets[\mathbf g_{n,1};\mathbf g_{n,2};\cdots;\mathbf g_{n,N_n}]^{T}$ \Comment{Initialize bag representation}
		\State $\mathbf{E}_n\gets$ the new bag representation by projecting $\mathbf G_n$ via a linear layer
		\State $\widehat{\mathbf{h}}_n\gets{{\arg \max}_{\mathbf{h}_n}}\ \mathbb{S}\left(\mathbb{D}\left(\mathbb{H}(\mathbf{E}_n,{\mathbf{h}}_n)\right)\right)$ \Comment{Generate indicator vector}
		\State $\widehat{\mathbf{E}}_n\gets \mathbb{H}(\mathbf{E}_n,\widehat{\mathbf{h}}_n)$ \Comment{Adaptively select representative instances within a bag}
		\State $\mathbf{B}_n\gets\mathbb{D}(\widehat{\mathbf{E}}_n)$ \Comment{Obtain sparse soft-bag representation}
		\State $\widehat{\mathbf T}_n\gets\mathbb{S}(\mathbf{B}_n)$ \Comment{Prognosis inference}
            \State
        $\underline{\mathbf{B}}$, $\widetilde{\mathbf{B}}_n\gets{\mathbb{K}_1}(\mathcal B^{\underline{\mathcal O}}),{\mathbb{K}_1}(\mathcal B_n^{\mathcal O})$  \Comment{Merge the representations of normal tissue and tumor tissue through $\mathbb{K}_1$}
		\State $\widehat{\mathbf{B}}_n\gets{\mathbb{K}_1}(\mathcal{B}_n^{\mathcal{O^+}}|\mathcal{B}_n^{\mathcal{O^-}})$ \Comment{Merge the bag representations (expect $\mathbf{B}_n$) from the same source with $\mathbf{B}_n$}
		\State $\overline{\mathbf{B}}_n\gets{\mathbb{K}_2}(\mathbb{H}(\mathbf{E}_n,\mathbf{1}-\widehat{\mathbf{h}}_n))$ \Comment{Merge the representation of the discarded instances}	
		\State $\mathcal{L}_{tcl}\gets$ the triple-tier contrastive learning loss calculated with $\{\mathbf{B}_n,\underline{\mathbf{B}},\widetilde{\mathbf{B}}_n,\widehat{\mathbf{B}}_n, \overline{\mathbf{B}}_n\}$
		\State $\mathcal{L}_{adc}\gets$ the absolute distance constraint calculated with $\{\mathbf{B}_n,\underline{\mathbf{B}},\widetilde{\mathbf{B}}_n,\widehat{\mathbf{B}}_n, \overline{\mathbf{B}}_n\}$
  		\State $\mathcal{L}_{cox}\gets$ the Cox loss  calculated with $\{\widehat{\mathbf T}_n,\mathbf T_n,\boldsymbol{\delta}_n\}$
		\EndFor
		\State $\mathcal L_{s}\gets$ the sparseness loss calculated with $\{\mathbf{W}_Q,\mathbf{W}_K,\mathbf{W}_V\}$
		\State $\mathcal L_{\textrm{II}}\gets\mathcal{L}_{cox}+\beta_{tcl}\mathcal{L}_{tcl}{ + \beta_{adc} \mathcal{L}_{adc}} + \beta_{s}\mathcal{L}_{s}$ \Comment{Hybrid loss of Curriculum II}
		\State Update $\{\mathbb{B},\mathbb{D},\mathbb{S}\}$ by gradient descent.
	\end{algorithmic}
\end{algorithm}

\begin{figure*}[t!]
    \renewcommand\thefigure{S\arabic{figure}}
    \centering
    \includegraphics[width=0.9\textwidth]{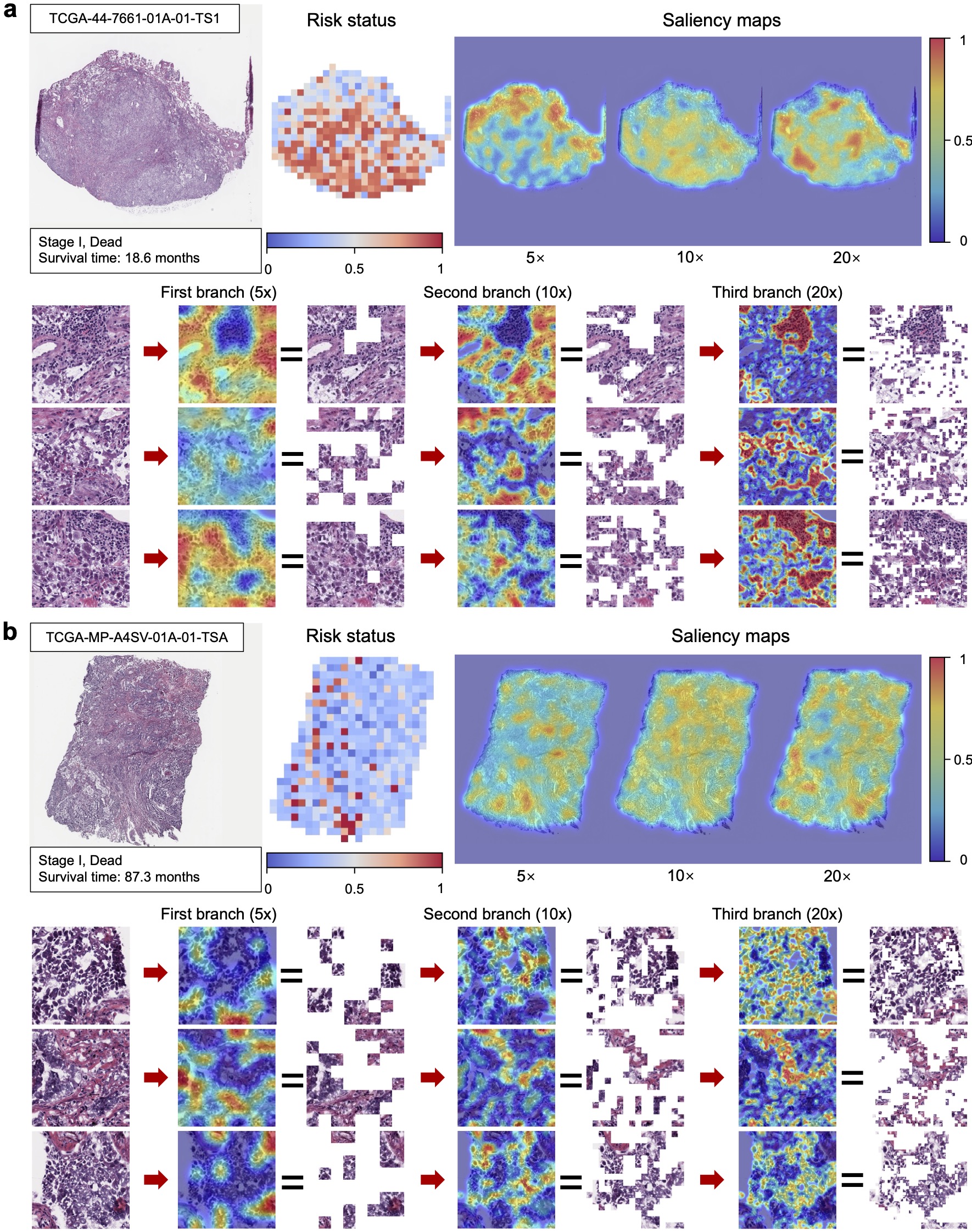}
    \caption{\textbf{Visualizations of saliency maps across multi-magnification WSIs on Curriculum I for LUAD dataset. Related to Figure 5.} This includes a representative high-risk case (\textbf{a}) which is stage I and deceased at 18.6 months, a representative low-risk case(\textbf{b}) which is stage I and deceased at 87.3 months. \textbf{a}, \textbf{b}, The first row shows original WSI, patch-level risk status estimation, and saliency maps of WSIs at different magnifications (i.e., 5$\times$, 10$\times$, and 20$\times$).
    The second row exhibits the outputs derived from the low-magnification branches, which guide fine-grained representation learning at higher magnifications.
    }	
    \label{fig_2_luad}
\end{figure*}

\begin{figure*}[t!]
    \renewcommand\thefigure{S\arabic{figure}}
    \centering
    \includegraphics[width=0.9\textwidth]{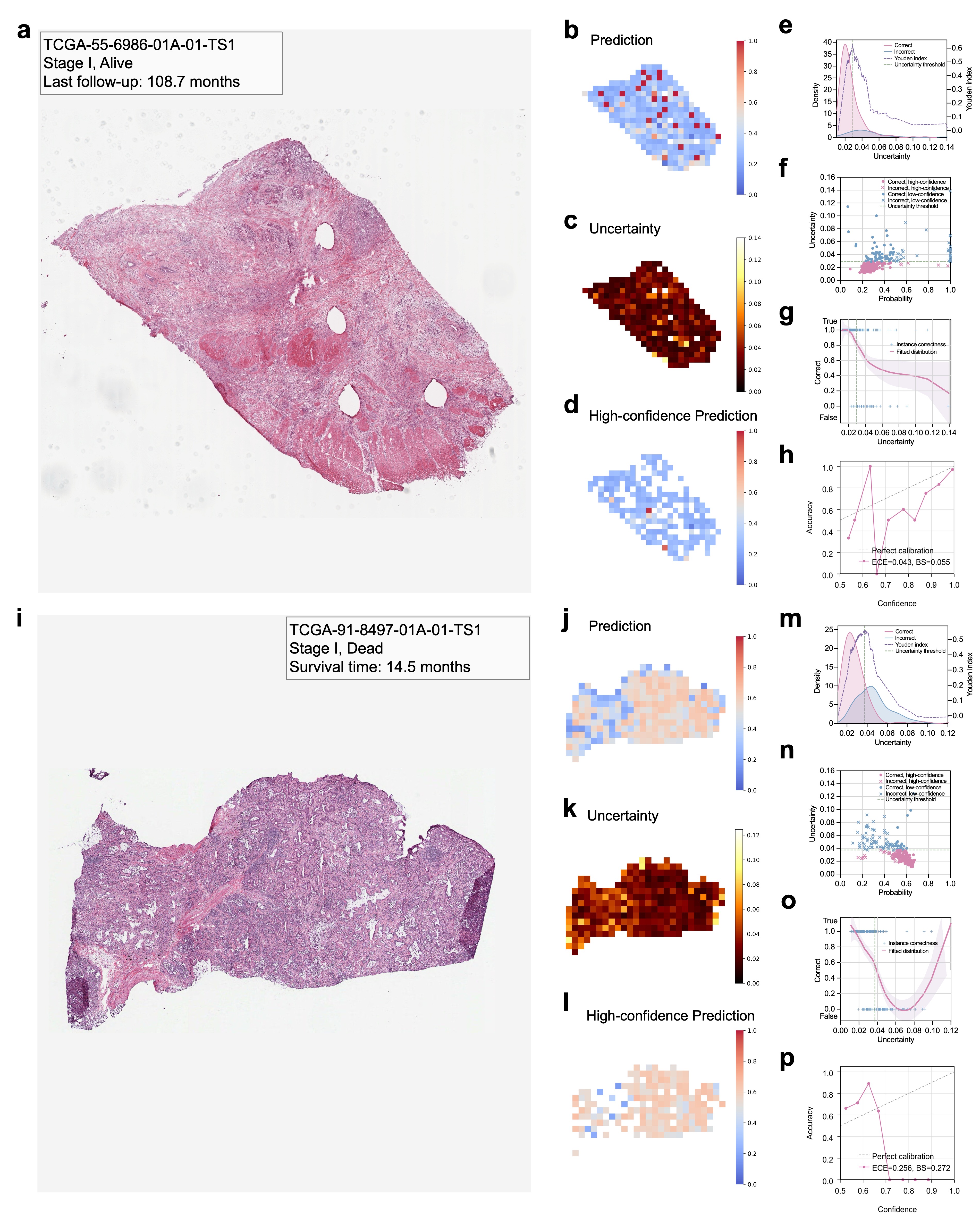}
    \caption{\textbf{Instance uncertainty estimation towards risk stratification on Curriculum I for LUAD dataset. Related to Figure 6.} It includes a high-risk case (stage I, deceased at 14.5 months) and a low-risk case (stage I, survived over 108.7 months).
    \textbf{a, i}, Original WSIs. \textbf{b, j}, Risk maps. \textbf{c, k}, Uncertainty maps. \textbf{d, l}, High-confidence instances. \textbf{e, m}, Probability density distributions (Red: correct; Blue: incorrect; Purple: Youden index). \textbf{f, n}, Instance uncertainty distributions. \textbf{g, o}, Correlation between uncertainty and prediction correctness. \textbf{h, p} show the reliability curves.}
    \label{fig_3_luad}
\end{figure*}

\begin{figure*}[t!]
    \renewcommand\thefigure{S\arabic{figure}}
    \centering
    \includegraphics[width=0.9\textwidth]{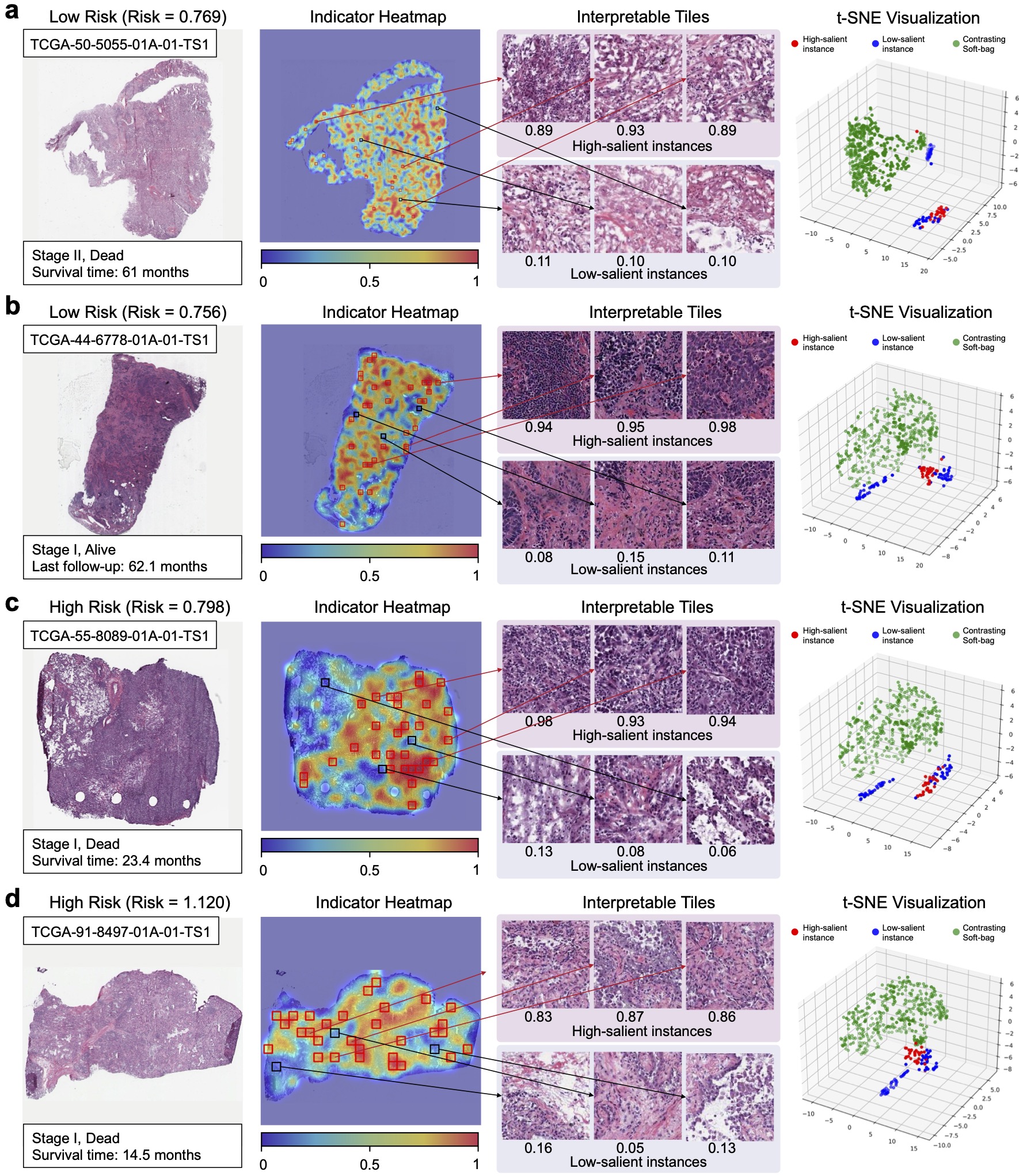}
    \caption{\textbf{Four representative cases from the COAD dataset are visualized on Curriculum II, illustrated with the original WSI, indicator heatmap, interpretable tiles, and t-SNE visualization. Related to Figure 6.} It includes two low-risk patients (\textbf{a}, stage II, deceased at 61 months; \textbf{b}, and stage I, survival times over 62.1 months) and two high-risk patients (\textbf{c}, stage I, deceased at 23.4 months; \textbf{d}, stage I, deceased at 14.5 months). \textbf{a}, \textbf{b}, \textbf{c}, \textbf{d}, The indicator heatmap visualizes the soft-bag instance selection and its localization within the WSI. Interpretive tiles exhibit several high-salient and low-salient instances, along with their corresponding indicator scores. The t-SNE visualization depicts the embedding distributions of high-salient instances, low-salient instances, and the contrasting soft-bags from other WSIs.}
    \label{fig_4_luad}
\end{figure*}

\begin{figure*}[t!]
    \renewcommand\thefigure{S\arabic{figure}}
    \centering
    \includegraphics[width=0.9\textwidth]{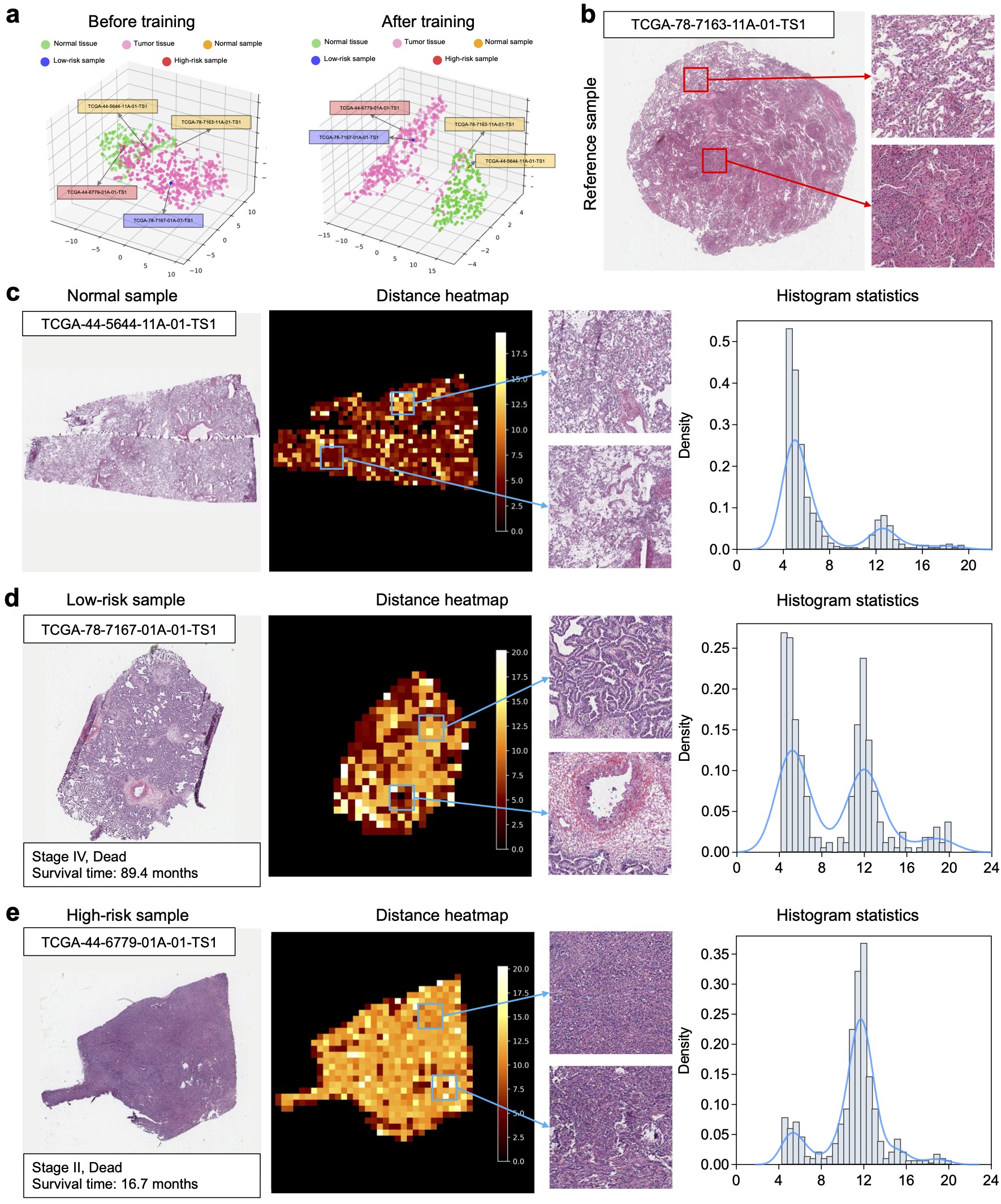}   \caption{\textbf{Visualization of comparative analysis between tumor and normal tissues on Curriculum II for LUAD dataset. Related to Figure S2.}
    \textbf{a}, t-SNE visualization shows the distribution of tumor and normal tissues before and after training. 
    \textbf{b}, a normal sample as reference.
    \textbf{c}, \textbf{d}, \textbf{e}, Three cases of comparisons are visualized, including a normal sample (\textbf{c}), a low-risk sample (\textbf{d}, stage IV, deceased at 89.4 months), and a high-risk sample (\textbf{e}, stage II, deceased at 16.7 months). The first column exhibits the original WSIs of these samples, the second column illustrates the distance heatmaps along with two representative tiles, and the third column shows the histogram statistics of the distance heatmaps.
    }
    \label{fig_6_luad}
\end{figure*}

\begin{figure*}[t!]
    \renewcommand\thefigure{S\arabic{figure}}
    \centering
    \includegraphics[width=0.9\textwidth]{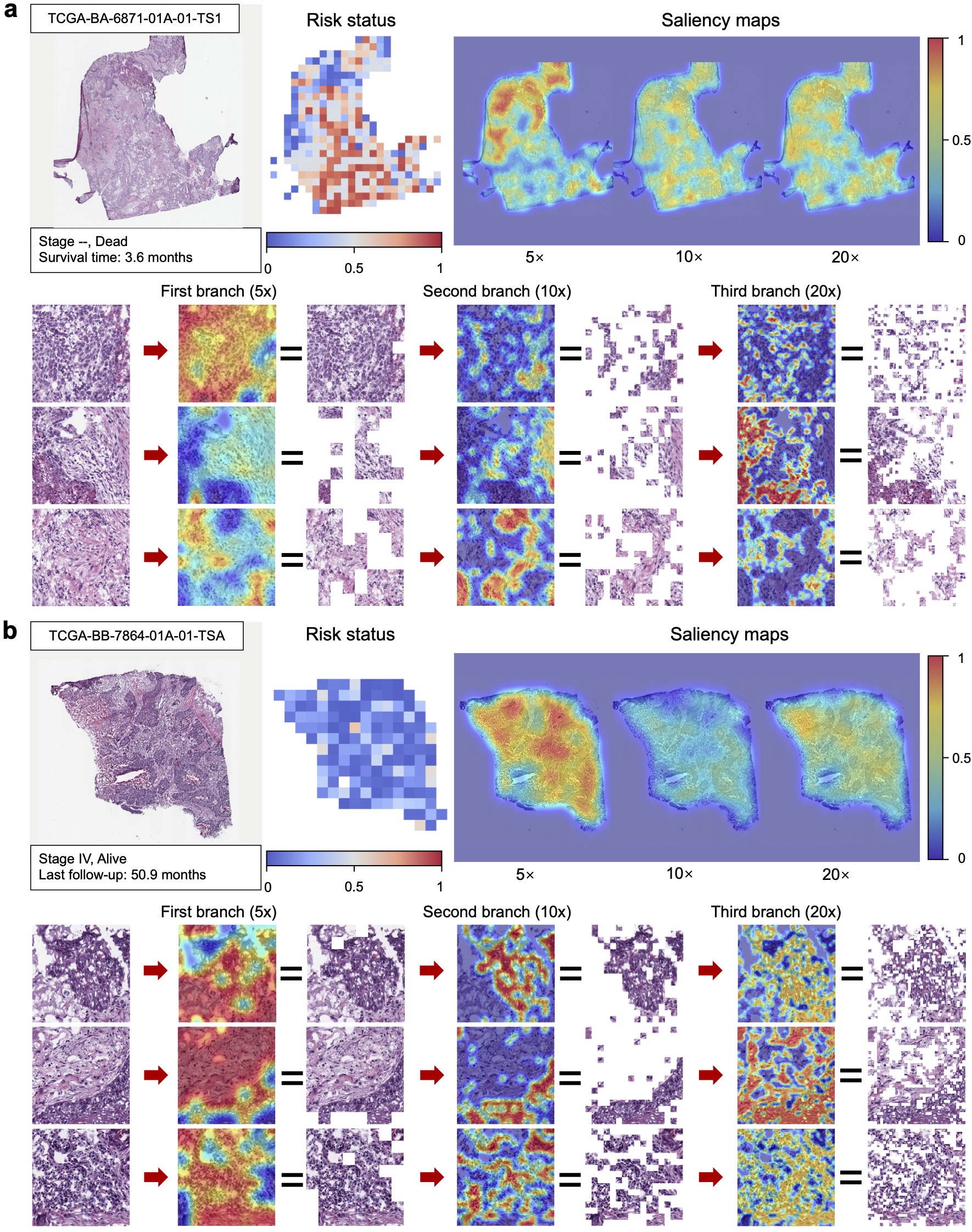}
    \caption{\textbf{Visualizations of saliency maps across multi-magnification WSIs on Curriculum I for HNSC dataset. Related to Figure 5.} This includes a representative high-risk case (\textbf{a}) which is stage -- and deceased at 3.6 months, a representative low-risk case(\textbf{b}) which is stage IV and survived over 50.9 months. \textbf{a}, \textbf{b}, The first row shows original WSI, patch-level risk status estimation, and saliency maps of WSIs at different magnifications (i.e., 5$\times$, 10$\times$, and 20$\times$). The second row exhibits the outputs derived from the low-magnification branches, which guide fine-grained representation learning at higher magnifications.
}
    \label{fig_2_hnsc}
\end{figure*}

\begin{figure*}[t!]
    \renewcommand\thefigure{S\arabic{figure}}
    \centering
    \includegraphics[width=0.9\textwidth]{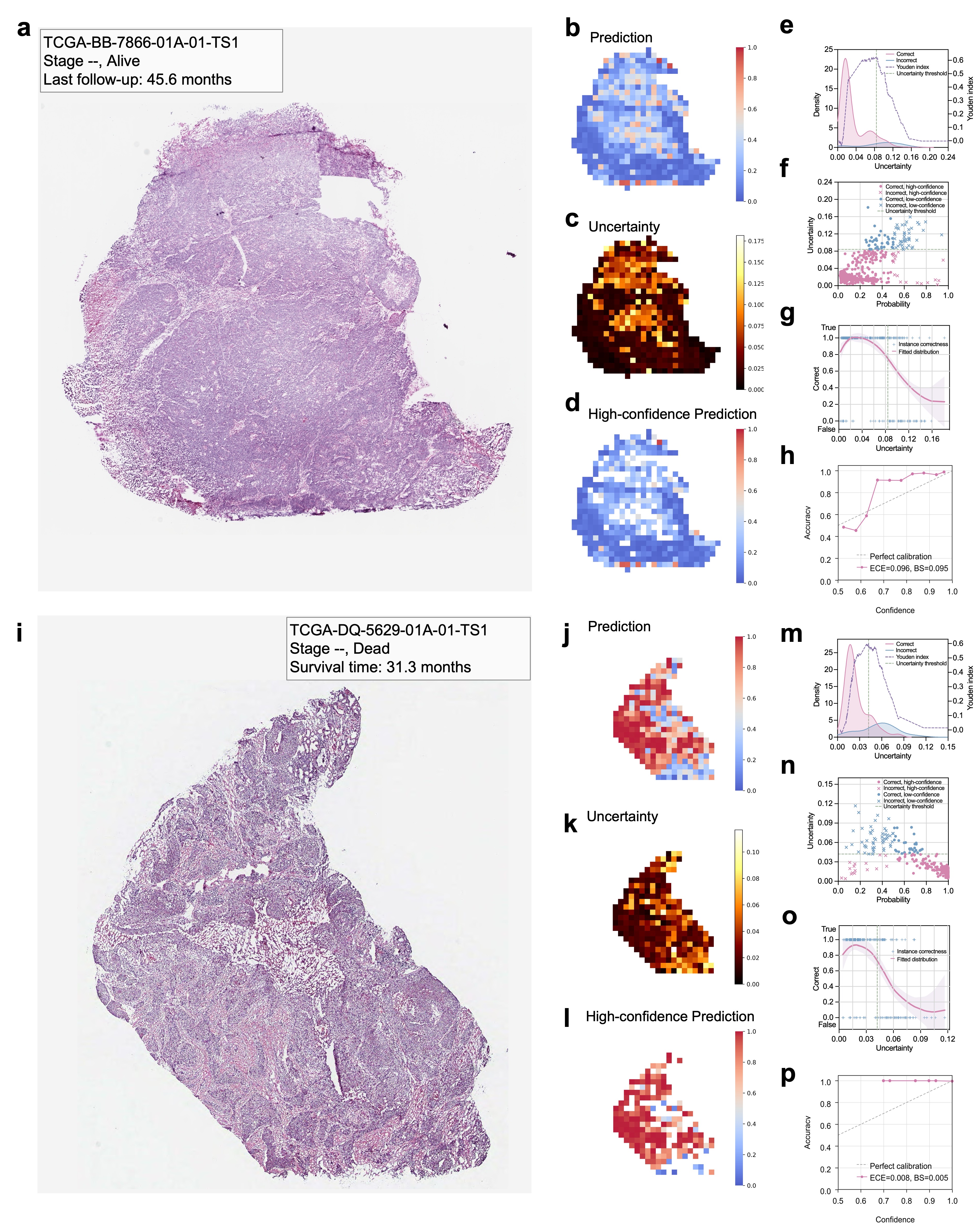}
    \caption{
    \textbf{Instance uncertainty estimation towards risk stratification on Curriculum I for HNSC dataset. Related to Figure 6.} It includes a high-risk case (stage --, deceased at 31.3 months) and a low-risk case (stage --, survived over 45.6 months).
    \textbf{a, i}, Original WSIs. \textbf{b, j}, Risk maps. \textbf{c, k}, Uncertainty maps. \textbf{d, l}, High-confidence instances. \textbf{e, m}, Probability density distributions (Red: correct; Blue: incorrect; Purple: Youden index). \textbf{f, n}, Instance uncertainty distributions. \textbf{g, o}, Correlation between uncertainty and prediction correctness. \textbf{h, p} show the reliability curves.}
    \label{fig_3_hnsc}
\end{figure*}

\begin{figure*}[t!]
    \renewcommand\thefigure{S\arabic{figure}}
    \centering
    \includegraphics[width=0.9\textwidth]{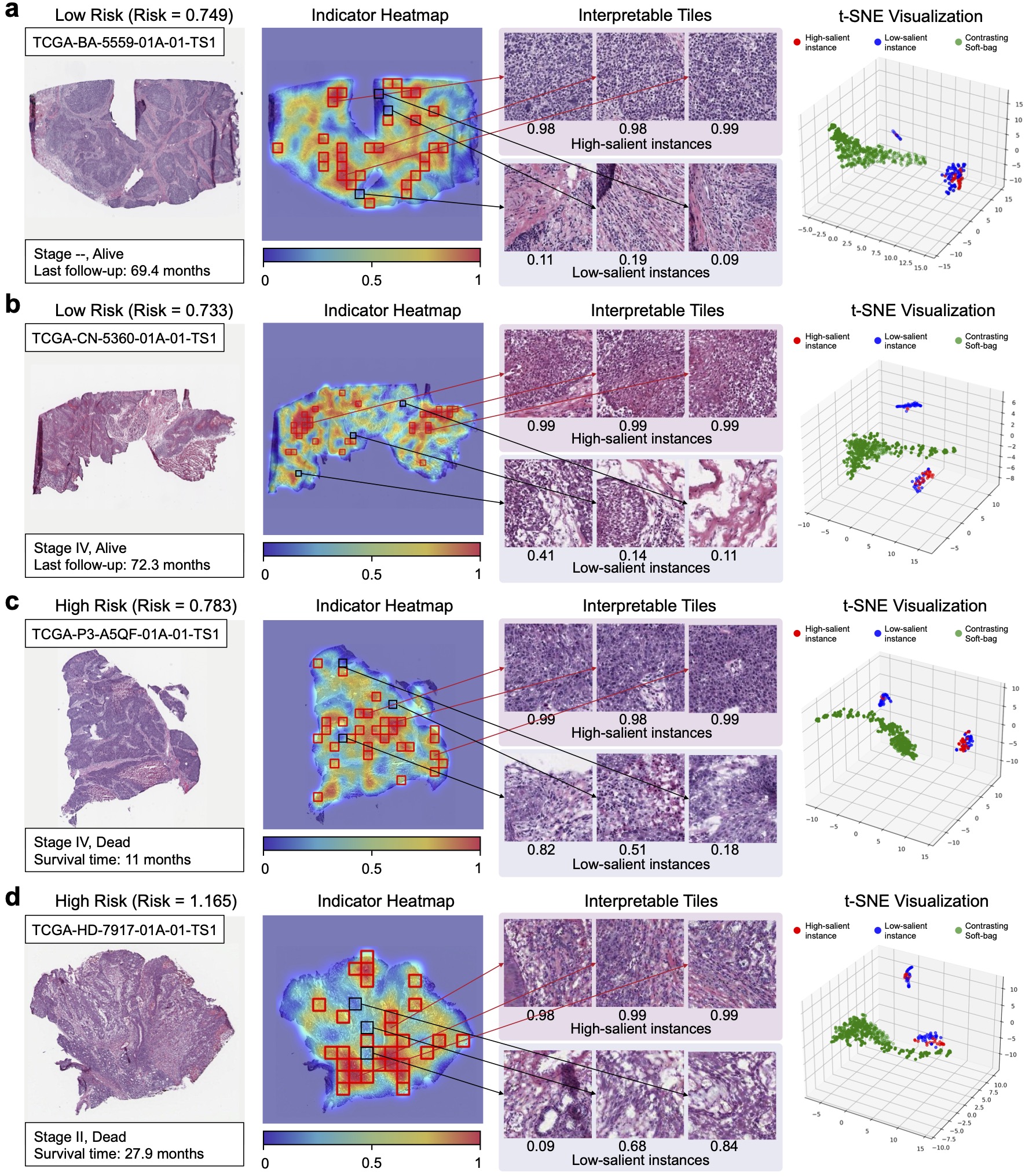}
    \caption{\textbf{Four representative cases from the HNSC dataset are visualized on Curriculum II, illustrated with the original WSI, indicator heatmap, interpretable tiles, and t-SNE visualization. Related to Figure 6.} It includes two low-risk patients (\textbf{a}, stage --, survival times over 69.4 months; \textbf{b}, and stage IV, survival times over 72.3 months) and two high-risk patients (\textbf{c}, stage IV, deceased at 11 months; \textbf{d}, stage II, deceased at 27.9 months). \textbf{a}, \textbf{b}, \textbf{c}, \textbf{d}, The indicator heatmap visualizes the soft-bag instance selection and its localization within the WSI. Interpretive tiles exhibit several high-salient and low-salient instances, along with their corresponding indicator scores. The t-SNE visualization depicts the embedding distributions of high-salient instances, low-salient instances, and the contrasting soft-bags from other WSIs.}
	\label{fig_4_hnsc}
\end{figure*}

\begin{figure*}[t!]
	\renewcommand\thefigure{S\arabic{figure}}
    \centering
    \includegraphics[width=0.9\textwidth]{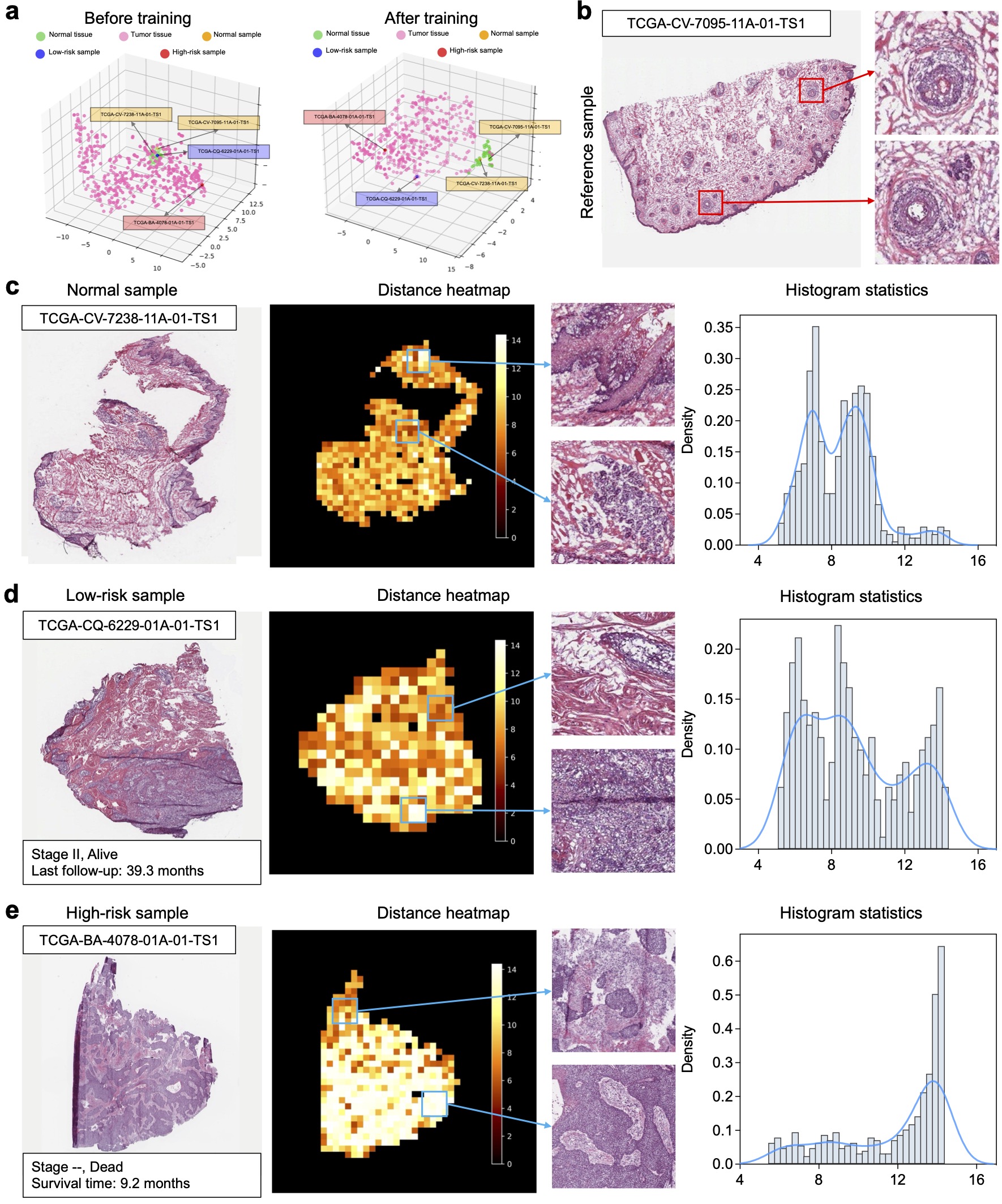}
    \caption{\textbf{Visualization of comparative analysis between tumor and normal tissues on Curriculum II for HNSC dataset. Related to Figure S2.}
    \textbf{a}, t-SNE visualization shows the distribution of tumor and normal tissues before and after training. 
    \textbf{b}, a normal sample as reference.
    \textbf{c}, \textbf{d}, \textbf{e}, Three cases of comparisons are visualized, including a normal sample (\textbf{c}), a low-risk sample (\textbf{d}, stage II, survival times over 39.3 months), and a high-risk sample (\textbf{e}, stage --, deceased at 9.2 months). The first column exhibits the original WSIs of these samples, the second column illustrates the distance heatmaps along with two representative tiles, and the third column shows the histogram statistics of the distance heatmaps.
    }
    \label{fig_6_hnsc}
\end{figure*}

\end{document}